\PassOptionsToPackage{table}{xcolor}
\documentclass{fairmeta}
\usepackage[utf8]{inputenc}
\usepackage{lmodern}

\usepackage{amsmath,amssymb,tabularx,array,colortbl,float}
\usepackage{url}
\definecolor{rapRankFirst}{HTML}{5EB95F}
\definecolor{rapRankSecond}{HTML}{A5DAA6}
\definecolor{rapRankThird}{HTML}{E4F8E4}
\DeclareRobustCommand{\RAP}{\texorpdfstring{{\rmfamily\bfseries\itshape R\scalebox{0.8}{AP}}}{RAP}}
\providecommand{\Description}[1]{}
\title{\RAP{}\kern0.10em: Research Attention Prediction in Language Agents}
\author[1]{Yingqian Wu}
\author[1]{Jingcong Liang}
\author[1]{Shengbin Yue}
\author[2]{Siyuan Wang}
\author[3]{Zhenfei Yin}
\author[3]{Junchi Yu}
\author[1,4]{Zhongyu Wei}
\affiliation[1]{Fudan University}
\affiliation[2]{The Chinese University of Hong Kong}
\affiliation[3]{University of Oxford}
\affiliation[4]{Shanghai Innovation Institute}

\hypersetup{pdfauthor={Yingqian Wu, Jingcong Liang, Shengbin Yue, Siyuan Wang, Zhenfei Yin, Junchi Yu, Zhongyu Wei},
pdftitle={RAP: Research Attention Prediction in Language Agents},
pdfkeywords={LLM agents, literature search, research attention prediction, rolling evaluation, benchmarks}}

\abstract{LLM agents increasingly support research planning and execution, where judgments about current and anticipated research activity inform decisions. Yet whether they can infer changes across a whole field from the partial evidence they retrieve through online scholarly search remains unclear. We introduce \textbf{Research Attention Prediction (\RAP{})}, a literature search benchmark for evaluating this ability. We build a local search environment from 356,357 dated arXiv papers and construct two complementary evaluation tasks. The benchmark comprises a Core set of 2,160 questions across 54 AI/ML fields and an Extended set covering 269 fields. Our evaluation and analysis of LLM agents show that search substantially improves stronger agents' recognition of recent changes, while forecasting gains remain limited or negative. Agents update forecasts across dates and adjust their investigations to forecasting goals, but these adaptations do not reliably yield accurate predictions of future shifts. Even for recent changes, final judgments can diverge from what the retrieved papers indicate. These findings show that agents can recognise some recent changes, yet reliably anticipating future research activity remains difficult despite changes in their answers and investigations.}

\metadata[Code and data]{\url{https://github.com/Win7win/rap-benchmark}}
\correspondence{Junchi Yu at \email{junchi.yu@eng.ox.ac.uk}, Zhongyu Wei at \email{zywei@fudan.edu.cn}}
\begin{document}
\maketitle

\begin{figure}[H]
\centering
\includegraphics[width=\textwidth]{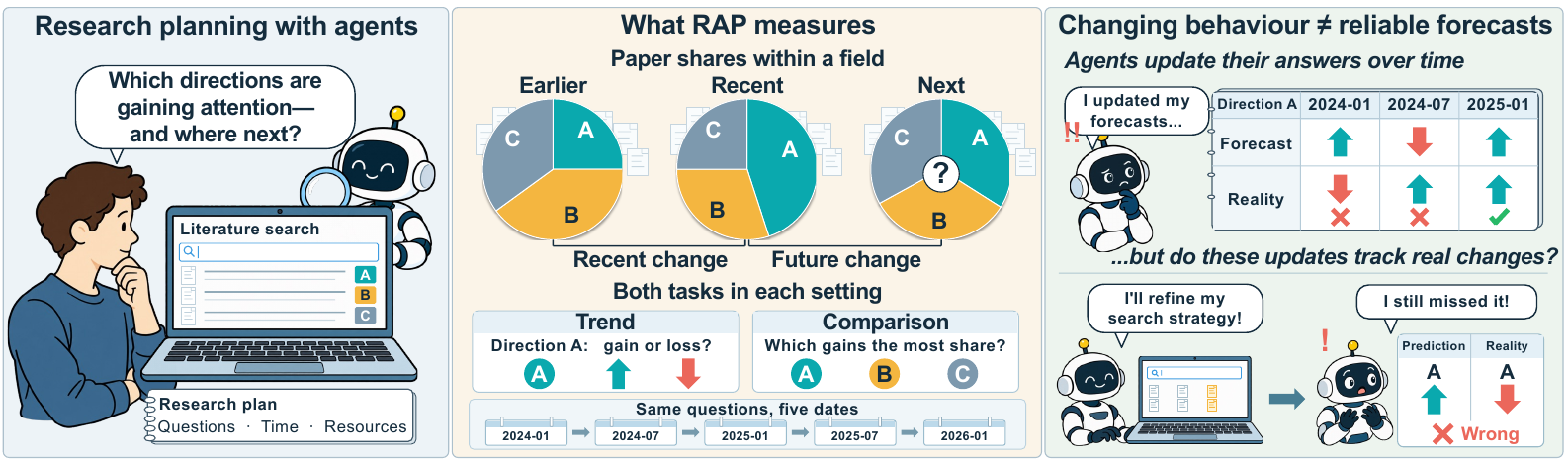}
\caption{\RAP{} evaluates whether LLM agents can infer recent and future changes in field-wide paper shares from partial literature evidence. Trend and Comparison tasks repeat at five dates. The illustrative scenes highlight two gaps: forecast updates need not track realised changes, and adjusting the search does not ensure reliable forecasts.}
\Description{Three columns connect research planning with agents, the RAP benchmark, and its findings. A researcher asks which directions are gaining attention and where attention will move next. Three pies illustrate earlier, recent, and future paper shares; Trend and Comparison tasks apply to both recent and future change and repeat over five dates. The upper-right scene compares changing forecasts for direction A with realised changes across dates. In the lower-right scene, an agent adjusts its search strategy but still produces a forecast that disagrees with reality. The pies and answer sequences are schematic illustrations, not measured results for specific questions.}
\label{fig:illustration}
\end{figure}

\begingroup
\newcommand{\BFields}{269}
\newcommand{\BDates}{five}
\newcommand{\BQuestions}{10{,}740}
\newcommand{\BPapers}{356{,}357}
\section{Introduction}

Recent progress in foundation models~\cite{openaigpt6sol2026,deepseekv41flash2026,qwen35modelcard2026} is reshaping the role of artificial intelligence (AI) in scientific research, from a passive conversational assistant to an active participant~\citep{coi2025}.
Beyond task execution, a greater challenge of scientific research lies in deciding which problems to pursue and how to allocate resources.
Such decisions take place within an evolving scientific landscape, where growing research effort does not necessarily lead to greater productivity~\citep{bloom2020ideas}, and increasing publication volumes have been associated with slower turnover of central scientific ideas~\citep{chu2021slowed}.
Understanding how research activity is distributed across scientific directions and how it may evolve can therefore provide valuable context for selecting research problems and guiding the allocation of research effort~\citep{rahkovsky2021portfolios,rafols2010overlay}.
Once a project begins, expectations about others' research activity can also shape how researchers allocate time and effort~\citep{hill2025competition}.

To inform these decisions, research agents use online scholarly search to form and update judgments about research activity~\citep{openscholar2024,coi2025}. Within practical time and computational budgets, they issue queries, analyse results and decide what to search next~\citep{foresci2026}. These search results provide a partial view of the literature, while the agents' judgments concern research activity across the whole field. We therefore ask: \emph{Can LLM agents retrieve relevant evidence and use it to infer recent and future shifts in the distribution of research activity within a field?}

Evaluating these search-based judgments raises three questions. \textbf{How to replay historical search environments?} Historical evaluation needs to assess what agents can infer at each date, with a search environment that recreates the corresponding literature access conditions. \textbf{How to distinguish forecasting failures from recognition failures?} Agents may recognise changes after they occur yet fail to predict them in advance~\citep{wu2026cusp}. Evaluating forecast errors alone does not reveal whether agents also struggle with recognition. \textbf{How to compare judgments as research activity evolves?} Plausible answers at individual dates do not show whether agents can track a changing research environment. Evaluation across dates should reveal whether agents update their judgments with evolving evidence and track changes in research activity.

To address these questions, we introduce \textbf{Research Attention Prediction (\RAP{})}, a literature search benchmark for identifying and predicting shifts in research activity. We build a local environment from \BPapers{} dated arXiv papers to simulate online scholarly search and measure research activity. We extract information from the papers to organise the corpus into fields, then group papers on similar research problems into directions within each field. For the \BFields{} resulting AI/ML fields, the environment provides interactive search that recreates literature access at each historical date. We define \emph{research attention} as the relative distribution of papers across directions within a field. By recording changes in research attention between successive six-month periods, we turn these shifts in research activity into evaluation tasks within the environment.

\RAP{} measures agents' ability to infer shifts in research activity across the whole field from the partial evidence they retrieve. We define two complementary tasks to assess individual trends and compare changes across directions. \textbf{Trend} asks whether a direction gains or loses share, and \textbf{Comparison} asks which candidate has the largest signed share change. To compare recognition and forecasting abilities, both tasks include paired retrospective \textbf{State} and prospective \textbf{Forecast} settings for the same changes. These paired tasks recur over \BDates{} successive semi-annual rounds with fixed questions and candidate directions, further testing whether judgments track actual changes in research activity. We apply this design to 54 fields to construct \textbf{Core}, with 2,160 questions for the main agent evaluation, while \textbf{Extended} broadens coverage to \BFields{} fields and \BQuestions{} questions.

Our experiments reveal a gap between agents' behavioural adaptation and their ability to anticipate shifts in research activity. \textbf{Search benefits recognition more than forecasting.} It substantially improves the stronger agents' State judgments, while forecasting gains remain limited or negative. \textbf{Answer changes do not imply temporal accuracy.} Agents update both State and Forecast answers across dates, but only State judgments reliably align with changes in their target periods. \textbf{Adapting investigations does not guarantee accurate judgments.} Agents tailor their investigations to forecasting goals, yet even in State, final judgments can diverge from changes reflected in the retrieved papers.
Overall, LLM agents can use retrieved evidence to detect recent shifts in research activity, but they struggle to anticipate future changes, even after updating predictions and adapting their investigations.

Our contributions are:
\begin{itemize}
    \item We introduce \RAP{}, extending literature search evaluation to recognising and predicting shifts in research activity. It assesses agents' ability to acquire relevant evidence and infer changes across a whole field from partial information.
    \item We build a literature environment that combines interactive search with measurements of research activity. By recreating literature access at different dates, it supports evaluation of agents' judgments against observed changes across a whole field.
    \item We reveal a gap between acquiring literature evidence and forming accurate judgments of research activity. Search helps stronger agents recognise recent changes, yet their judgments can miss changes reflected in retrieved papers and their forecast updates do not reliably track actual shifts.
\end{itemize}
\endgroup

\begin{figure*}[t]
\centering
\includegraphics[width=0.98\textwidth]{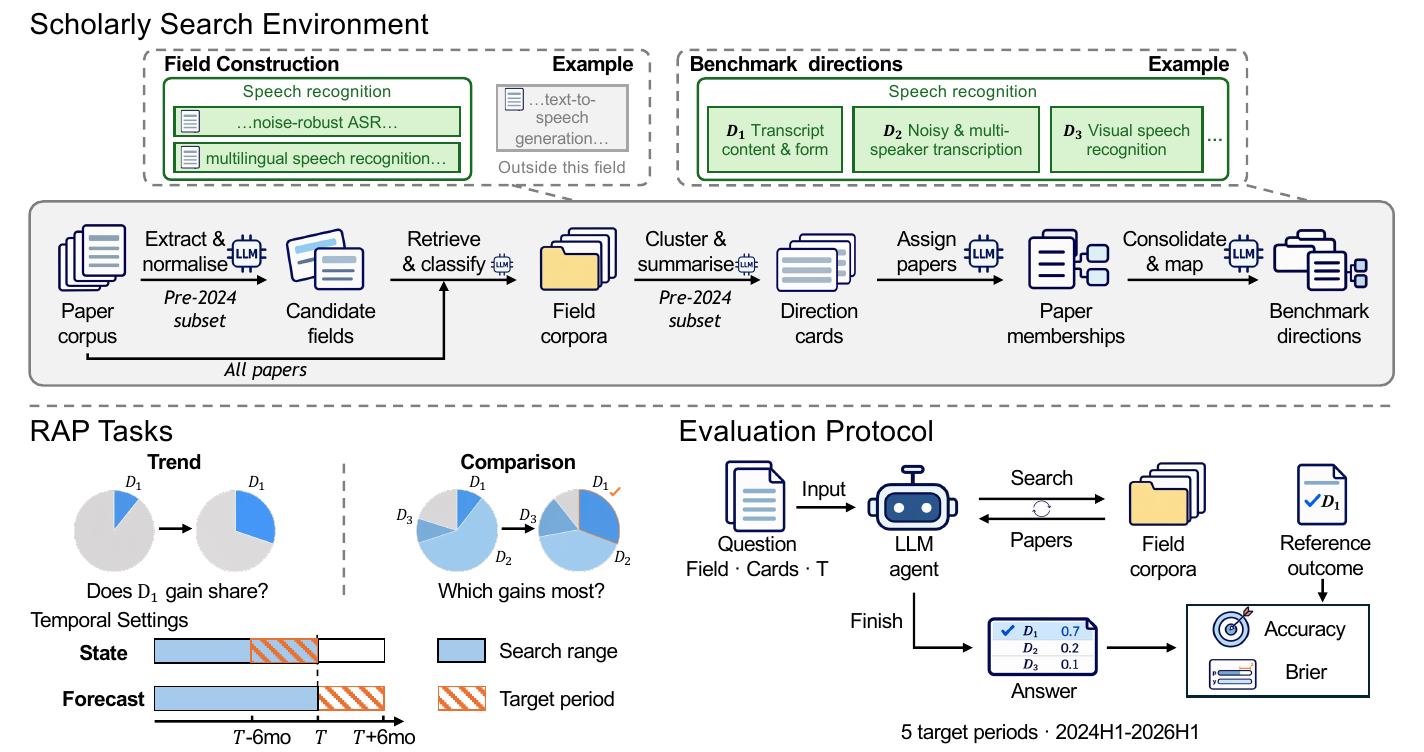}
\caption{Overview of \RAP{}. We organise papers into fields and benchmark directions to construct the scholarly search environment. Building on this environment, \RAP{} defines two tasks, Trend and Comparison, each with two temporal settings, State and Forecast. Agents search pre-cut-off literature and submit answers for scoring across five semi-annual target periods.}
\Description{The upper pipeline extracts and normalises field labels from pre-2024 papers, retrieves and classifies papers into field corpora, constructs direction cards from historical papers, assigns paper memberships, and consolidates benchmark directions. Two speech-recognition examples illustrate field boundaries and final directions. Below, paired pies show a single direction gaining share and compare share changes across candidate directions. State compares the latest observed half-year with its predecessor; Forecast compares the next half-year with the latest observed one. Both restrict search to papers before cut-off T. On the right, an agent receives a question, field, direction cards and cut-off, searches field corpora and returns answer probabilities. The probabilities and reference outcome feed accuracy and Brier scoring. Five target periods span 2024 H1 to 2026 H1.}
\label{fig:rap-pipeline}
\end{figure*}

\section{Related Work}

\paragraph{Evaluating literature-based LLM agents.}
Scholarly search helps researchers develop an understanding of a topic through iterative investigation~\citep{marchionini2006exploratory}. LLM agents apply this process to scientific question answering, literature synthesis and research ideation~\citep{openscholar2024,paperqa2024,researchagent2025,coi2025}. Beyond literature-specific applications, SMART learns coordinated agent trajectories for knowledge-intensive tasks~\citep{yue2025synergistic}, while ParaAgent learns to organise tool exploration and execution through parallel actions~\citep{yue2026paraagent}. Existing evaluations of LLM agents examine both evidence acquisition and the quality of their outputs. SAGE studies how agents' queries interact with their retrieval tools~\citep{sage2026}. ScholarQABench, DeepResearch Bench and ResearcherBench assess correctness, coverage and evidence support in scientific answers, research reports and advice~\citep{openscholar2024,deepresearchbench2025,researcherbench2025}. \RAP{} complements these evaluations by checking judgments about shifts in research attention against observed changes in publication shares.

\paragraph{Evaluating scientific judgment in LLMs.}
Existing benchmarks evaluate LLMs' scientific judgments against observed research outcomes. BrainBench evaluates predictions of experimental results, and CUSP assesses predictions of scientific advances and compares performance with time-restricted and unrestricted Web search~\citep{luo2025brainbench,wu2026cusp}. PreScience covers both paper-level outcomes and aggregate topic-share changes~\citep{prescience2026}. ForeSci evaluates direction selection and research planning using cutoff-aligned literature and validation against later papers~\citep{foresci2026}. \citet{tong2026ailearnscientifictaste} study learning scientific judgment and idea generation from community feedback, including citation-based preferences. \RAP{} examines LLMs' scientific judgment through changes in research attention. It retains interactive retrieval and date-restricted evidence access, adding paired recognition and prediction of the same changes and rolling evaluation of fixed questions. Table~\ref{tab:benchmark-comparison} summarises these evaluation features across benchmarks.

\begin{table}[t]
\caption{Comparison of benchmark features. Search: interactive retrieval; Cutoff: date-restricted evidence; Paired: forecasting and recognition of the same outcomes; Rolling: fixed targets across successive evidence and outcome windows. $\checkmark$: included; $\circ$: selected tasks or diagnostics; ---: not evaluated.}
\label{tab:benchmark-comparison}
\centering\small
\setlength{\tabcolsep}{2.5pt}
\begin{tabularx}{\columnwidth}{@{}>{\raggedright\arraybackslash}Xcccc@{}}
\toprule
\textbf{Benchmark} & \textbf{Search} & \textbf{Cutoff} & \textbf{Paired} & \textbf{Rolling} \\
\midrule
ScholarQABench~\citep{openscholar2024} & $\checkmark$ & --- & --- & --- \\
SAGE~\citep{sage2026} & $\checkmark$ & --- & --- & --- \\
DeepResearch Bench~\citep{deepresearchbench2025} & $\checkmark$ & --- & --- & --- \\
\midrule
PreScience~\citep{prescience2026} & $\circ$ & $\checkmark$ & --- & --- \\
CUSP~\citep{wu2026cusp} & $\circ$ & $\circ$ & $\circ$ & --- \\
ForeSci~\citep{foresci2026} & $\checkmark$ & $\checkmark$ & --- & --- \\
\midrule
\textbf{\RAP{}} & $\checkmark$ & $\checkmark$ & $\checkmark$ & $\checkmark$ \\
\bottomrule
\end{tabularx}
\end{table}

\section{The \RAP{} Benchmark}
\label{sec:benchmark}

\RAP{} evaluates LLM agents' judgments of recent and future changes in research activity in an interactive literature-search environment. Figure~\ref{fig:rap-pipeline} outlines the environment, tasks and evaluation.

\subsection{A Time-Aware Scholarly Search Environment}
\label{sec:environment}

To simulate online scholarly search with controlled literature access, we build a local environment from an arXiv paper corpus. Agents use the partial evidence they retrieve to infer recent and future shifts in the distribution of research activity across the whole field.

\paragraph{Construction.}
We collect 356,357 arXiv papers first submitted between June 2022 and June 2026. From the pre-January 2024 papers, we extract research descriptors with Qwen3.8-27B~\citep{qwen38modelcard2026} using complete titles and abstracts. We then normalise the field labels in these descriptors with GPT-6-sol~\citep{openaigpt6sol2026} and retain candidate fields supported by at least 50 construction papers. Candidate retrieval followed by Qwen3.8-27B classification expands their paper memberships. Consolidating remaining name variants yields 271 overlapping \emph{research fields}, such as speech recognition and image restoration. Within each field, we group papers by embedding similarity and construct an intermediate set of fine-grained direction cards with GPT-6-sol. Each card specifies a direction's name, scope, inclusion and exclusion criteria, and supporting papers. Qwen3.8-27B assigns papers to the cards using complete titles and abstracts.

To align the granularity of directions across fields, we further merge related fine-grained groups into broader research-problem families. We use few-shot prompting~\citep{brown2020fewshot} to guide this consolidation, allowing at most 15 directions per field without prescribing an exact count. Paper memberships are then mapped from the fine-grained groups to the final \emph{benchmark directions} for activity measurement. Excluding two single-direction fields leaves 269 benchmark fields and 2,187 directions, averaging 8.1 directions per field (range 2--15). Appendix~\ref{app:construction} details construction and membership rules.
\paragraph{Reference validity.}
Two researchers and GPT-6-sol assess the same 100 papers across four fields (Table~\ref{tab:human-model-agreement}). Human--model agreement varies across reviewers, while the two models agree more often. The criterion measures overlap between assigned direction sets, rather than exact-set agreement or assignment accuracy. This distinction matters for aggregation: adding a second membership halves a paper's weight in its original direction under fractional counting. Switching to full membership counts changes 9.1\% of Trend references and 13.0\% of Comparison answer sets on jointly scorable Core events. State search gains nevertheless remain larger than Forecast gains under this alternative and under coverage and margin filters (Appendix~\ref{app:reference-robustness}).

\begin{table}[t]
\centering\small
\caption{Direction-set overlap (\%) on 100 papers: 25 each in LLM reasoning, information extraction, parameter-efficient fine-tuning and LLM alignment. Sets agree if they intersect or both are empty.}
\label{tab:human-model-agreement}
\begin{tabular}{@{}lrr@{}}
\toprule
Researcher & Qwen3.8-27B & GPT-6-sol \\
\midrule
A & 78 & 77 \\
B & 63 & 62 \\
\midrule
Qwen--Sol & \multicolumn{2}{c}{92} \\
\bottomrule
\end{tabular}
\end{table}

\paragraph{Search access.}
We provide a field-local \textsc{Search} tool using BM25, a keyword-based ranking function. The tool accepts keyword queries, optional date bounds, and a choice of relevance or date ordering. For a cut-off $T$, it searches only papers assigned to the question's field and first submitted before $T$. It returns dates, complete titles and abstracts, and matching-record counts. The interface does not expose paper--direction assignments or reference activity counts.

\subsection{\RAP{} Tasks}
\label{sec:tasks}

\paragraph{Research attention.}
For a research field $f$, a benchmark direction $d$ within it, and a time window $W$, let $s_{f,d}(W)$ denote the direction's share of covered papers. Each paper contributes unit mass, divided equally among its confirmed benchmark directions within the field. Papers without a confirmed direction are excluded from the denominator. Shares are normalised over the full field. Appendix~\ref{app:references} gives the aggregation formula and reference-handling rules.
At an evaluation cut-off $T$, we define three six-month windows: $W_{-1}=[T-12\mathrm{mo},T-6\mathrm{mo})$ is the earlier observed half-year, $W_0=[T-6\mathrm{mo},T)$ is the latest observed half-year, and $W_1=[T,T+6\mathrm{mo})$ is the next half-year. Here $\mathrm{mo}$ denotes months; each interval includes its start date and excludes its end date. The corresponding signed changes in share are
\begin{equation}
 \begin{aligned}
 \Delta^{\mathrm{S}}_{f,d}(T)&=s_{f,d}(W_0)-s_{f,d}(W_{-1}),\\
 \Delta^{\mathrm{F}}_{f,d}(T)&=s_{f,d}(W_1)-s_{f,d}(W_0).
 \end{aligned}
\label{eq:attention-change}
\end{equation}
The superscripts $\mathrm{S}$ and $\mathrm{F}$ denote \RAP{}'s two temporal settings, \textsc{State} and \textsc{Forecast}, respectively. \textsc{State} takes a retrospective view, reconstructing the change between the two observed half-years. \textsc{Forecast} takes a prospective view, predicting the change in the next half-year before papers in $W_1$ become available. Both use the same pre-$T$ search permissions, without access to exact shares.

\paragraph{Task definitions.}
Both temporal settings support two tasks, with $\Delta$ denoting the corresponding share change defined above. A \emph{Trend} question asks whether one direction's share increases or decreases between the specified half-years. A \emph{Comparison} question asks which direction in a candidate set $C_f$ of benchmark directions from field $f$ has the largest signed change in field-wide share. We compare percentage-point changes, not final share levels or percentage growth rates. If all candidates decline, the smallest decline wins. Example questions appear in Appendix~\ref{app:task-examples}.

\subsection{Evaluation Protocol}
\label{sec:rolling-protocol}

\paragraph{Agent interaction.}
Each episode provides a field, question, direction cards and cut-off $T$. In the search-enabled Agent setting, the LLM formulates queries and examines retrieved evidence within a fixed search budget. It then submits probabilities over the possible answers through a \textsc{Finish} tool. See Appendix~\ref{app:answers} for output rules and Appendix~\ref{app:full-inputs} for complete prompts and tool specifications.

\paragraph{Rolling evaluation.}
Following rolling-origin evaluation~\citep{hyndman2021forecasting}, we evaluate agents at six-month intervals to examine how their judgments change as new literature becomes available. Each question starts an independent episode. The main evaluation uses five Forecast cut-offs from January 2024 through January 2026 and five State cut-offs from July 2024 through July 2026. Forecast($T$) and State($T+6$ months) concern the same realised change but differ in available evidence. Appendix~\ref{app:rolling} provides the alignment details.

\paragraph{Sample construction.}
We sample up to three directions per field without using reference answers or model scores, yielding one Trend question per sampled direction and one Comparison question over the same candidates. Directions, definitions and candidate order remain fixed across temporal targets and rolling cut-offs. We sample 54 fields from the environment, stratified by the number of directions per field, and construct the Core set with 2,160 questions. Applying the same question-construction rules to all 269 benchmark fields produces the Extended set with 10,740 questions (Table~\ref{tab:rolling-inventory}). Sampling details are given in Appendix~\ref{app:rolling}.

\begin{table}[t]
\caption{Benchmark size over five rolling rounds per temporal target. Core is a subset of Extended.}
\label{tab:rolling-inventory}
\centering
\small
\begin{tabular}{@{}lrr@{}}
\toprule
\textbf{Inventory} & \textbf{Core} & \textbf{Extended} \\
\midrule
Benchmark fields & 54 & 269 \\
Benchmark directions & 453 & 2,187 \\
Directions per field (mean) & 8.4 & 8.1 \\
Sampled directions & 162 & 805 \\
Rounds per target & 5 & 5 \\
Questions & 2,160 & 10,740 \\
\bottomrule
\end{tabular}
\end{table}

\section{Main Results}
\label{sec:results}

\subsection{Experimental Setup}

\paragraph{Metrics.}
\label{sec:metrics}

We report field-macro accuracy and output validity (Valid), averaging within fields and then equally across fields. Invalid answers count as incorrect. Brier loss~\citep{brier1950verification} measures probability quality; lower is better. See Appendix~\ref{app:evaluation-details} for scoring details.

\paragraph{Models and inference.}
We evaluate eleven models: Llama-3.2-3B-Instruct~\citep{llama32modelcard2024} and Llama-3.1-8B-Instruct~\citep{grattafiori2024llama3}; Qwen3-4B and Qwen3-8B~\citep{qwen3technicalreport2025}, Qwen3.5-4B~\citep{qwen35modelcard2026} and Qwen3.8-27B~\citep{qwen38modelcard2026}; gpt-oss-120b~\citep{openaiossmodelcard2025}; DeepSeek-V4.1-Flash~\citep{deepseekv41technicalreport2026}; and GPT-5.6-terra~\citep{openaigpt56systemcard2026}, GPT-6-luna and GPT-6-sol~\citep{openaigpt6systemcard2026} (Table~\ref{tab:core-main}). Open models use native templates. Thinking is enabled for Qwen, GPT, DeepSeek and gpt-oss models, with reasoning effort set to high where supported and otherwise left at its default. We request temperature 1, with a 64K-token context and a 16K-token output limit per call; DeepSeek's thinking API does not apply temperature. Agent allows up to 20 searches, each returning up to eight complete titles and abstracts; Closed answers without search. Both receive the same questions and selected directions' names and scopes. Core scores average three independent runs for every model and setting; Extended Closed evaluation uses one complete run. Provider-specific handling and run-level variation are reported in Appendix~\ref{app:evaluation-details}.

\paragraph{Baselines.}
Uniform choice gives the chance-level reference. A fixed-search control uses predetermined queries based on the field and candidate names, assembles their results, and obtains one GPT-6-sol answer per question in each run. It shares Agent's corpus, temporal cut-off, search allowance, full-abstract returns and context limit, without adapting queries to retrieved evidence. Neither workflow receives evaluator memberships or exact historical shares. Query construction is detailed in Appendix~\ref{app:evaluation-details}.

\subsection{Core-set Evaluation}

Table~\ref{tab:core-main} reports model performance on the Core set under Closed (without search) and Agent (with search) settings.

\begin{table*}[!t]
\caption{Core field-macro accuracy (\%) over five same-event pairs, averaged over three independent runs for each model and setting. Uniform choice is the analytic expectation. Within each Closed/Agent accuracy column, dark-to-light green marks the top three models; the best is bold.}
\label{tab:core-main}
\newcommand{\rapClosedRows}{%
Llama-3.2-3B-Instruct & 48.15 & 24.69 & 48.52 & 25.68 & 93.5 \\
Llama-3.1-8B-Instruct & 48.97 & 26.91 & 49.34 & 31.11 & 96.8 \\
Qwen3-4B & 47.98 & 35.06 & 47.98 & 32.84 & 100.0 \\
Qwen3-8B & 48.44 & 31.11 & 48.19 & 30.86 & 99.9 \\
Qwen3.5-4B & 48.19 & 32.35 & 45.27 & 30.25 & 93.0 \\
Qwen3.8-27B & 49.26 & 30.25 & 49.51 & 32.22 & 96.4 \\
gpt-oss-120b & 49.96 & 29.75 & 49.84 & 30.99 & 100.0 \\
DeepSeek-V4.1-Flash & 49.30 & 10.49 & 49.47 & 10.62 & 82.6 \\
GPT-5.6-terra & \cellcolor{rapRankFirst}\textbf{52.47} & \cellcolor{rapRankThird}36.17 & \cellcolor{rapRankSecond}52.43 & \cellcolor{rapRankThird}35.19 & 100.0 \\
GPT-6-luna & \cellcolor{rapRankSecond}52.43 & \cellcolor{rapRankSecond}37.16 & \cellcolor{rapRankThird}51.65 & \cellcolor{rapRankSecond}36.67 & 100.0 \\
GPT-6-sol & \cellcolor{rapRankThird}52.30 & \cellcolor{rapRankFirst}\textbf{39.63} & \cellcolor{rapRankFirst}\textbf{53.05} & \cellcolor{rapRankFirst}\textbf{37.90} & 100.0 \\
}

\newcommand{\rapAgentRows}{%
Llama-3.2-3B-Instruct & 45.97 & 19.26 & 48.02 & 25.93 & 85.4 \\
Llama-3.1-8B-Instruct & 37.04 & 33.95 & 41.93 & 32.59 & 82.1 \\
Qwen3-4B & 1.89 & 13.21 & 8.48 & 17.04 & 19.5 \\
Qwen3-8B & 19.71 & 34.94 & 24.73 & 28.15 & 56.3 \\
Qwen3.5-4B & 50.62 & 32.10 & \cellcolor{rapRankThird}48.81 & 32.10 & 98.0 \\
Qwen3.8-27B & \cellcolor{rapRankThird}63.58 & \cellcolor{rapRankThird}50.37 & 47.82 & 32.84 & 99.8 \\
gpt-oss-120b & 55.02 & 41.98 & 46.87 & 31.85 & 99.3 \\
DeepSeek-V4.1-Flash & \cellcolor{rapRankSecond}64.12 & \cellcolor{rapRankSecond}52.47 & 45.35 & 32.35 & 97.7 \\
GPT-5.6-terra & 61.48 & 41.48 & \cellcolor{rapRankSecond}49.14 & \cellcolor{rapRankFirst}\textbf{35.43} & 100.0 \\
GPT-6-luna & 55.88 & 39.75 & 48.27 & \cellcolor{rapRankSecond}35.06 & 100.0 \\
GPT-6-sol & \cellcolor{rapRankFirst}\textbf{72.59} & \cellcolor{rapRankFirst}\textbf{56.67} & \cellcolor{rapRankFirst}\textbf{52.67} & \cellcolor{rapRankThird}34.69 & 100.0 \\
}

\newcommand{\rapFixedSearchRows}{%
GPT-6-sol (fixed search) & 69.79 & 58.40 & 51.52 & 33.33 & 100.0 \\
}

\centering\small
\setlength{\tabcolsep}{8pt}
\begin{tabular}{@{}lrrrrr@{}}
\toprule
& \multicolumn{2}{c}{\textbf{State}} & \multicolumn{2}{c}{\textbf{Forecast}} & \\
\cmidrule(lr){2-3}\cmidrule(lr){4-5}
\textbf{Model / method} & \textbf{Trend} & \textbf{Comparison} & \textbf{Trend} & \textbf{Comparison} & \textbf{Valid} \\
\midrule
\multicolumn{6}{@{}c@{}}{\textit{Closed (no search)}} \\
\addlinespace[2pt]
Uniform choice (expected) & 50.00 & 33.33 & 50.00 & 33.33 & 100.0 \\
\addlinespace[2pt]
\rapClosedRows
\midrule
\multicolumn{6}{@{}c@{}}{\textit{Fixed-search control}} \\
\addlinespace[2pt]
\rapFixedSearchRows
\midrule
\multicolumn{6}{@{}c@{}}{\textit{Agent (with search)}} \\
\addlinespace[2pt]
\rapAgentRows
\bottomrule
\end{tabular}
\end{table*}

\paragraph{Search helps recover changes in research attention.}
GPT-6-sol, Qwen3.8-27B, gpt-oss-120b, DeepSeek-V4.1-Flash and GPT-5.6-terra improve over Closed on both State tasks. GPT-6-sol leads the agents, gaining 20.3 percentage points on State Trend. GPT-6-luna shows smaller gains. Both Llama agents and Qwen3-4B/8B lose State Trend accuracy with search, with mixed effects on Comparison; Qwen3.5-4B remains near uniform-choice accuracy on both State tasks. Fixed search also gives Sol strong State performance, comparable to Agent on Comparison. Literature evidence thus supports recovery of recent changes without requiring every query to be chosen adaptively.

\paragraph{Forecasting remains a limitation for the agents.}
Models that improve on both State tasks nevertheless lose accuracy on Forecast Trend when given search access. Their Comparison gains are mostly small or absent; DeepSeek's larger gain coincides with much higher output validity (Appendix~\ref{app:execution-analysis}). The same State--Forecast contrast appears in Sol's fixed-search control, which improves both State tasks but lowers accuracy on both Forecast tasks relative to Closed. The paired results reveal a pronounced dependence on task setting for the same target events: with search, these models recognise realised changes much more accurately than they predict them, whereas the gap is much smaller without search.

For GPT-6-sol, Qwen3.8-27B, DeepSeek-V4.1-Flash and gpt-oss-120b, State search gains remain larger than Forecast gains after excluding questions with near-zero share changes or nearly tied candidate changes. The contrast also holds in a separate check restricted to events where at least 90\% of field papers receive a confirmed direction in each half-year (Appendix~\ref{app:boundary-coverage}). A fixed-history candidate-order check yields similar aggregate Comparison accuracy for GPT-6-sol (Appendix~\ref{app:candidate-order}).

\subsection{Extended-Set Evaluation}

Table~\ref{tab:closed-full} reports Closed performance on the Extended set.

\begin{table}[t]
\caption{Extended-set Closed field-macro accuracy (\%) over five same-event pairs, using one complete run per model. Bold indicates the best model accuracy.}
\label{tab:closed-full}
\newcommand{\rapFullRows}{%
Llama-3.2-3B-Instruct & 49.12 & 24.39 & 50.13 & 27.21 \\
Llama-3.1-8B-Instruct & 47.76 & 28.48 & 50.26 & 30.71 \\
Qwen3-4B & 48.97 & 34.94 & 49.67 & 35.61 \\
Qwen3-8B & 48.90 & 33.61 & 49.30 & 36.88 \\
Qwen3.5-4B & 48.04 & 30.93 & 46.70 & 34.50 \\
Qwen3.8-27B & 48.77 & 33.61 & 50.15 & 35.84 \\
gpt-oss-120b & 51.01 & 34.20 & 49.99 & 35.76 \\
DeepSeek-V4.1-Flash & 49.62 & 10.93 & 50.19 & 12.12 \\
GPT-5.6-terra & 51.10 & 36.58 & 51.93 & 37.10 \\
GPT-6-luna & 51.51 & 38.07 & 51.40 & \textbf{39.78} \\
GPT-6-sol & \textbf{51.72} & \textbf{38.36} & \textbf{52.47} & 38.88 \\
}

\centering\small
\setlength{\tabcolsep}{3pt}
\begin{tabular}{@{}lcccc@{}}
\toprule
& \multicolumn{2}{c}{\textbf{State}} & \multicolumn{2}{c}{\textbf{Forecast}} \\
\cmidrule(lr){2-3}\cmidrule(lr){4-5}
\textbf{Closed model} & \textbf{Trend} & \textbf{Comparison} & \textbf{Trend} & \textbf{Comparison} \\
\midrule
\rapFullRows
\bottomrule
\end{tabular}
\end{table}

\begin{table*}[!t]
\caption{Core Agent Forecast accuracy (\%) across all rounds and wholly post-release target windows, for models with eligible windows and using the same runs as Table~\ref{tab:core-main}. $\Delta$ is post-release minus all-round accuracy in percentage points.}
\label{tab:knowledge-coverage}
\newcommand{\rapTemporalResultRows}{%
Llama-3.2-3B-Instruct & 48.02 & 47.81 & -0.22 & 25.93 & 24.49 & -1.44 \\
Llama-3.1-8B-Instruct & 41.93 & 41.70 & -0.23 & 32.59 & 32.30 & -0.29 \\
Qwen3-4B & 8.48 & 10.80 & +2.33 & 17.04 & 19.44 & +2.41 \\
Qwen3-8B & 24.73 & 24.59 & -0.14 & 28.15 & 30.25 & +2.10 \\
gpt-oss-120b & 46.87 & 42.80 & -4.07 & 31.85 & 30.86 & -0.99 \\
}

\centering\small
\setlength{\tabcolsep}{9pt}
\begin{tabular}{@{}lrrrrrr@{}}
\toprule
& \multicolumn{3}{c}{\textbf{Trend}} & \multicolumn{3}{c}{\textbf{Comparison}} \\
\cmidrule(lr){2-4}\cmidrule(lr){5-7}
\textbf{Core Agent} & \textbf{All} & \textbf{Post-release} & $\boldsymbol{\Delta}$ & \textbf{All} & \textbf{Post-release} & $\boldsymbol{\Delta}$ \\
\midrule
\rapTemporalResultRows
\bottomrule
\end{tabular}
\end{table*}

Without search, State and Forecast accuracies remain similar across the broader field inventory. All three GPT models' Trend accuracies differ by less than one percentage point between the tasks, and their Comparison scores are also close. Open models remain near uniform-choice accuracy on most cells. The large State--Forecast gap for the strongest agents is absent from their Closed results, both on Core and across the Extended set.

\subsection{Post-release Forecast Evaluation}
\label{sec:temporal-coverage}

We further examine Core Agent forecasts in target windows wholly after model release, using the dates documented in Appendix~\ref{app:temporal-audit}. All eligible models score below uniform choice on both tasks (Table~\ref{tab:knowledge-coverage}). Relative to their all-round results, gpt-oss-120b shows the largest further decline on Trend, losing 4.1 percentage points. The Llama agents change little; Qwen3's Comparison scores rise slightly.

\section{Analysis}
\label{sec:analysis}

Section~\ref{sec:results} shows a marked State--Forecast gap, particularly in the benefit agents obtain from search. What explains this difference? We examine GPT-6-sol, Qwen3.8-27B, DeepSeek-V4.1-Flash and gpt-oss-120b from two perspectives: whether their judgments follow changes across time, and how their investigations and use of evidence differ between State and Forecast.

\subsection{Do Judgments Track the Time of Change?}
\label{sec:rolling-alignment}

\RAP{} asks the same questions at five successive dates, keeping the field and candidate directions fixed. This lets us examine two distinct properties: whether agents change their answers over time, and whether these changes track shifts in research attention.

\paragraph{Both temporal settings elicit changing answers.}
Across three complete Agent runs, all four models show greater answer variation between adjacent dates than between repeated runs at the same date, on both tasks (Appendix~\ref{app:rolling-alignment}). State answers change more frequently across dates than Forecast answers.

\begin{figure}[!htbp]
\centering
\includegraphics[width=0.66\linewidth]{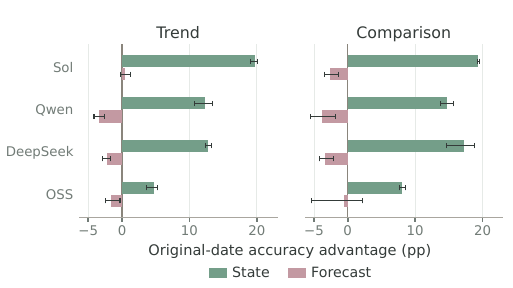}
\caption{Original-date accuracy advantage over expected permutation. Bars average three runs; whiskers show their range. Positive values indicate temporal alignment.}
\Description{Two horizontal grouped-bar panels show Trend and Comparison. Sage bars denote State and dusty rose bars Forecast. State advantages are positive for every model and task; Forecast advantages are near zero for Sol Trend and negative for the remaining combinations. Thin whiskers show the range of three runs.}
\label{fig:date-alignment}
\end{figure}

\paragraph{Temporal alignment is stronger for State than Forecast.}
To compare temporal alignment in State and Forecast, we compare original accuracy with its expectation under uniform permutation of each question's five answers across its five target periods. This preserves the question, answer frequencies and realised outcomes, removing only their temporal correspondence. It tests whether agents place their answers at the correct dates, independently of their overall preference for a candidate direction or for gain over loss. State accuracy falls under this reassignment for every model and task. Forecast accuracy is almost unchanged or improves (Figure~\ref{fig:date-alignment}): Sol's Comparison accuracy, for example, rises from 34.7\% to 37.3\%. Forecast answers thus vary without reliably identifying when the predicted change occurs.

\paragraph{Probability updates exhibit the same contrast.}
The contrast extends beyond the most likely answer. For Forecast, reassigning probability vectors lowers Brier loss in every model and task except Sol on Trend, where the original alignment retains a small benefit. State probabilities consistently fit their original periods better. A decomposition of Trend Brier loss further shows weaker discrimination between future gains and losses in all four models (Appendix~\ref{app:calibration-details}). Discrimination measures the improvement in Brier loss achieved by calibrated probabilities over a constant event-frequency prediction. Even after monotone calibration, Forecast probabilities offer little such improvement: their confidence estimates provide limited separation between future gains and losses. The temporal accuracy pattern also persists under three alternative membership-counting policies (Appendix~\ref{app:rolling-alignment}).

Rolling evaluation reveals a specific forecasting weakness: agents vary their predictions across dates, but neither the selected answers nor the assigned probabilities reliably identify when research attention will shift.

\subsection{How Do Agents Investigate and~Use~Evidence?}
\label{sec:same-cutoff}

We next investigate why agents gain less from search in Forecast than in State, examining how task goals shape their investigations and how they use the resulting evidence. We begin with State, where the observed change can be measured in both the retrieved sample and the full corpus, and then test whether its investigation histories and assessments can help improve Forecast.

\paragraph{Retrieved composition and final judgments can diverge.}
For all four models' State trajectories across three Agent runs, we compute the answer implied by the papers still visible at the final turn using their frozen memberships, then compare it with the full-corpus reference and the model's judgment. On matched questions, the retrieved-paper answer agrees with the reference more often for every model on both tasks (Table~\ref{tab:retrieved-composition}); the Trend gap is 4.5--12.6 percentage points. The retrieved sample already contains a useful aggregate signal, but the final judgment does not fully recover it. Discrepancies therefore arise both in acquiring evidence representative of the field and in interpreting the evidence obtained. Appendix~\ref{app:visible-composition} details the calculation.

\begin{table}[!htbp]
\centering
\caption{Agreement with the full-corpus State reference (\%). Retrieved denotes the answer computed from visible papers using frozen memberships; Model denotes the final judgment. Each pair uses matched eligible questions, averaged over three runs.}
\label{tab:retrieved-composition}
\small
\setlength{\tabcolsep}{3pt}
\begin{tabular}{@{}lrrrr@{}}
\toprule
& \multicolumn{2}{c}{Trend} & \multicolumn{2}{c}{Comparison} \\
\cmidrule(lr){2-3}\cmidrule(l){4-5}
Model & Retrieved & Model & Retrieved & Model \\
\midrule
GPT-6-sol & 77.0 & 72.6 & 60.0 & 57.0 \\
Qwen3.8-27B & 76.2 & 63.6 & 61.1 & 50.7 \\
DeepSeek-V4.1-Flash & 75.8 & 64.6 & 61.7 & 52.7 \\
gpt-oss-120b & 66.5 & 56.2 & 46.1 & 42.5 \\
\bottomrule
\end{tabular}
\end{table}

\begin{figure*}[t]
\centering
\includegraphics[width=\textwidth]{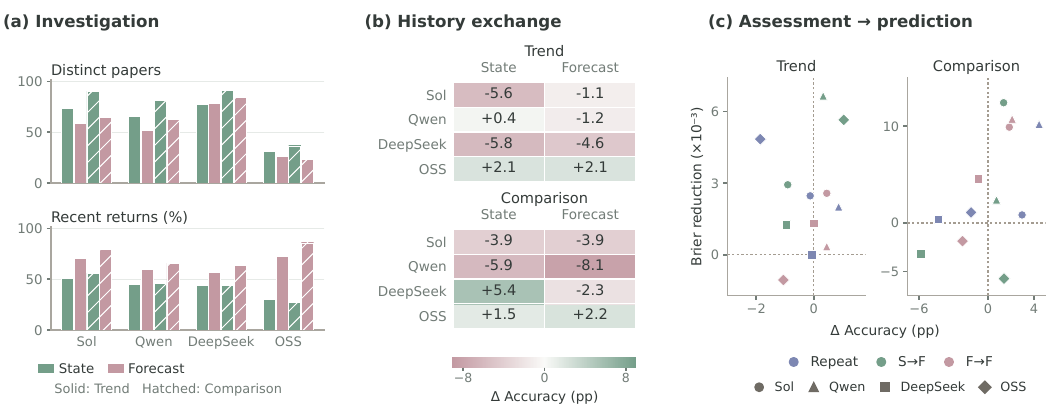}
\caption{Investigation and evidence use. \textbf{(a)} Distinct papers and recent returns over three runs; recent counts include duplicates. Sage denotes State, rose Forecast; solid bars show Trend, hatched bars Comparison. \textbf{(b)} Accuracy change (pp) from exchanging histories, relative to the mean of two own-history judgments; columns identify the final task. \textbf{(c)} Accuracy change and Brier reduction relative to the original Forecast; right and up indicate improvement. Continuations average three draws for every model; original and final-step-repeat answers remain fixed. DeepSeek and OSS use shared-valid questions in (b,c), matched within each draw in (c). Appendix~\ref{app:analysis-details} details the protocols and output-validity sensitivity.}
\Description{Three horizontally aligned panels combine grouped bars, signed effect matrices and accuracy-versus-Brier scatter plots. Panel a shows more recent retrieval under Forecast for all four models. Panel b shows that history exchange reduces Forecast accuracy for Sol, Qwen and DeepSeek and increases it for OSS; sage indicates accuracy gains and dusty rose losses. Panel c compares repeat judgments, State-then-Forecast and Forecast-then-Forecast on Trend and Comparison, retaining the original scores and matching rules. Smoky blue is reserved for the repeat control in panel c.}
\label{fig:investigation-evidence}
\label{fig:search-goals}
\label{fig:history-transfer}
\label{fig:forecast-continuation}
\end{figure*}

We next compare investigations at the same origin, with identical search access and budgets. Forecast concentrates more returns in the preceding six months for all four models; Sol, Qwen and gpt-oss also retrieve fewer distinct papers (Figure~\ref{fig:investigation-evidence}a). This recency contrast is consistent with the tasks' temporal requirements: State compares two observed half-years; Forecast projects into the future.

\paragraph{Exchanging investigation histories.}
Would the different investigation conducted for State provide a better basis for forecasting? For all four models, we exchange the complete State and Forecast histories at the same cut-off, retaining the target question and regenerating only its final judgment. State histories reduce Forecast accuracy for Sol, Qwen and DeepSeek on both tasks (Figure~\ref{fig:investigation-evidence}b). gpt-oss instead benefits on shared-valid questions, although its Trend Brier loss does not improve. These histories retain queries, assistant text, tool returns and omission markers; the contrast tests the acquired investigation as a whole. The effect of changing the investigation depends on how each model uses the resulting history for prediction.

\paragraph{Separating assessment from prediction.}
We next test whether explicitly assessing the observed change helps agents use existing evidence for prediction. We keep the original Forecast question and history fixed, asking the model to produce a same-cut-off State assessment before the final forecast (S$\rightarrow$F). Forecast reassessment followed by Forecast (F$\rightarrow$F) provides a two-call control; a final-step repeat measures re-answering variation. Across three continuation draws per model, S$\rightarrow$F has lower mean accuracy than F$\rightarrow$F for Sol and DeepSeek on both tasks (Figure~\ref{fig:investigation-evidence}c). Qwen's Trend accuracies are nearly tied, while Comparison favours F$\rightarrow$F. Compared with F$\rightarrow$F, S$\rightarrow$F lowers Brier loss for Sol on both tasks and for Qwen on Trend. For gpt-oss, mean accuracy improves over the original Forecast on shared-valid questions, but the gain reverses in the third draw on both tasks, and Comparison Brier loss worsens. Including incomplete answers removes both mean gains (Table~\ref{tab:diagnostic-validity}).

\paragraph{Sol is the most active reviser when forecasting.}
Across the four models, we score the intermediate State answer against its own period and track how it is revised for the future. DeepSeek and gpt-oss have mean intermediate-answer retention of at least 92\% on valid pairs, including frequent carry-forward when State is correct but the future reference differs. Qwen revises more, and Sol revises most (Appendix~\ref{app:state-forecast-transition}). Sol correctly identifies 66.5\% of observed Trend changes; its subsequent revisions improve future accuracy over carrying the State answer forward by eight to nine points across the two tasks. These revisions add predictive value to the intermediate answer, but final accuracy remains below F$\rightarrow$F. Explicitly separating the steps exposes both the quality of the State assessment and the value added by revising it for the future.

The rolling and paired analyses reveal a gap between adapting to the task and anticipating its target change. Agents update forecasts across dates and alter their investigations, but these adaptations do not reliably identify when research attention will shift. Their intermediate judgments further expose different ways of carrying observed changes forward or revising them for the future, with limited additional predictive value.

\section{Discussion and Limitations}
\label{sec:limitations}

\paragraph{Implications for LLM agents.}
The results highlight a gap between acquiring literature and using it to judge changes in research attention. For State, final judgments can diverge from changes reflected in the retrieved papers; for Forecast, answers change across dates without reliably tracking the corresponding periods. A central challenge for LLM agents is therefore to turn partial literature evidence into accurate judgments of how research attention changes and when those changes occur.

\paragraph{Rolling evaluation as diagnosis.}
\RAP{} separates two questions that a single forecasting score leaves unresolved: can an agent anticipate a shift, and can it recognise that shift once the corresponding literature is available? Rolling evaluation adds a further check of whether changes in answers follow changes in research activity. Errors that persist after publication motivate work on reconstructing activity from literature alongside better forecasting methods. Assessing reported confidence together with accuracy can distinguish justified certainty from stronger but incorrect judgments.

\paragraph{Limitations.}
\RAP{} measures relative publication activity in arXiv, with agent evaluation limited to field-local BM25 search. Its findings require further validation under open-Web search. Model-built memberships and counting choices introduce reference uncertainty, and counting robustness does not establish semantic correctness (Appendix~\ref{app:reference-robustness}). Retrieval cut-offs do not remove later information in revised abstracts or model parameters. Behavioural analyses use the Core subset, while recurring questions and overlapping fields limit independence across observations.

\section{Conclusion}

We introduced \RAP{}, a time-indexed literature-search environment and rolling benchmark for judging recent and future changes in research attention, measured by within-field paper shares. Search substantially improves stronger agents' recognition of realised changes, but does not consistently improve their forecasts. Rolling evaluation reveals that forecasts vary across dates without reliably tracking when attention shifts. Paired evaluation shows that agents adapt evidence acquisition to forecasting goals under identical literature access, yet these adaptations do not reliably produce more accurate predictions. Together, these findings identify a gap between adapting an investigation to its temporal goal and using the resulting evidence to anticipate field-level changes. \RAP{} makes this gap measurable and tests whether better search and reasoning improve judgments as research activity evolves.

\bibliographystyle{assets/plainnat}
\bibliography{refs}

\begin{thebibliography}{43}
\providecommand{\natexlab}[1]{#1}
\providecommand{\url}[1]{\texttt{#1}}
\expandafter\ifx\csname urlstyle\endcsname\relax
  \providecommand{\doi}[1]{doi: #1}\else
  \providecommand{\doi}{doi: \begingroup \urlstyle{rm}\Url}\fi

\bibitem[Ajith et~al.(2026)Ajith, Singh, DeYoung, Kunievsky, Kozlowski, Tafjord, Evans, Weld, Hope, and Downey]{prescience2026}
Anirudh Ajith, Amanpreet Singh, Jay DeYoung, Nadav Kunievsky, Austin~C. Kozlowski, Oyvind Tafjord, James Evans, Daniel~S. Weld, Tom Hope, and Doug Downey.
\newblock {PreScience}: A dataset and benchmark for scientific forecasting, 2026.
\newblock \url{https://arxiv.org/abs/2602.20459}.

\bibitem[Asai et~al.(2026)Asai, He, Shao, et~al.]{openscholar2024}
Akari Asai, Jacqueline He, Rulin Shao, et~al.
\newblock Synthesizing scientific literature with retrieval-augmented language models.
\newblock \emph{Nature}, 650\penalty0 (8103):\penalty0 857--863, 2026.

\bibitem[Baek et~al.(2025)Baek, Jauhar, Cucerzan, and Hwang]{researchagent2025}
Jinheon Baek, Sujay~Kumar Jauhar, Silviu Cucerzan, and Sung~Ju Hwang.
\newblock {ResearchAgent}: Iterative research idea generation over scientific literature with large language models.
\newblock In \emph{Proceedings of the 2025 Conference of the Nations of the Americas Chapter of the Association for Computational Linguistics: Human Language Technologies (Volume 1: Long Papers)}, pages 6709--6738, Albuquerque, New Mexico, 2025. Association for Computational Linguistics.
\newblock \doi{10.18653/v1/2025.naacl-long.342}.

\bibitem[Bloom et~al.(2020)Bloom, Jones, Van~Reenen, and Webb]{bloom2020ideas}
Nicholas Bloom, Charles~I. Jones, John Van~Reenen, and Michael Webb.
\newblock Are ideas getting harder to find?
\newblock \emph{American Economic Review}, 110\penalty0 (4):\penalty0 1104--1144, 2020.
\newblock \doi{10.1257/aer.20180338}.

\bibitem[Brier(1950)]{brier1950verification}
Glenn~W. Brier.
\newblock Verification of forecasts expressed in terms of probability.
\newblock \emph{Monthly Weather Review}, 78\penalty0 (1):\penalty0 1--3, 1950.
\newblock \doi{10.1175/1520-0493(1950)078<0001:VOFEIT>2.0.CO;2}.

\bibitem[Brown et~al.(2020)Brown, Mann, Ryder, Subbiah, Kaplan, Dhariwal, Neelakantan, Shyam, Sastry, Askell, Agarwal, Herbert-Voss, Krueger, Henighan, Child, Ramesh, Ziegler, Wu, Winter, Hesse, Chen, Sigler, Litwin, Gray, Chess, Clark, Berner, McCandlish, Radford, Sutskever, and Amodei]{brown2020fewshot}
Tom Brown, Benjamin Mann, Nick Ryder, Melanie Subbiah, Jared~D Kaplan, Prafulla Dhariwal, Arvind Neelakantan, Pranav Shyam, Girish Sastry, Amanda Askell, Sandhini Agarwal, Ariel Herbert-Voss, Gretchen Krueger, Tom Henighan, Rewon Child, Aditya Ramesh, Daniel Ziegler, Jeffrey Wu, Clemens Winter, Chris Hesse, Mark Chen, Eric Sigler, Mateusz Litwin, Scott Gray, Benjamin Chess, Jack Clark, Christopher Berner, Sam McCandlish, Alec Radford, Ilya Sutskever, and Dario Amodei.
\newblock Language models are few-shot learners.
\newblock In H.~Larochelle, M.~Ranzato, R.~Hadsell, M.~F. Balcan, and H.~Lin, editors, \emph{Advances in Neural Information Processing Systems}, volume~33, pages 1877--1901, Red Hook, NY, USA, 2020. Curran Associates, Inc.
\newblock \url{https://proceedings.neurips.cc/paper/2020/hash/1457c0d6bfcb4967418bfb8ac142f64a-Abstract.html}.

\bibitem[Chu and Evans(2021)]{chu2021slowed}
Johan S.~G. Chu and James~A. Evans.
\newblock Slowed canonical progress in large fields of science.
\newblock \emph{Proceedings of the National Academy of Sciences}, 118\penalty0 (41):\penalty0 e2021636118, 2021.
\newblock \doi{10.1073/pnas.2021636118}.

\bibitem[{DeepSeek-AI}(2026{\natexlab{a}})]{deepseekv41flash2026}
{DeepSeek-AI}.
\newblock {DeepSeek-V4.1-Flash}: Smarter, faster, more efficient, September 2026{\natexlab{a}}.
\newblock \url{https://api-docs.deepseek.com/news/news260910/}.
\newblock Published 10 September 2026; accessed 26 October 2026.

\bibitem[{DeepSeek-AI}(2026{\natexlab{b}})]{deepseekv41technicalreport2026}
{DeepSeek-AI}.
\newblock {DeepSeek-V4.1-Flash}: Pushing the limits of {KV} cache compression.
\newblock Technical report, DeepSeek-AI, 2026{\natexlab{b}}.
\newblock \url{https://huggingface.co/deepseek-ai/DeepSeek-V4.1-Flash/resolve/main/DeepSeek_V41_Tech_Report.pdf}.

\bibitem[Dimitriadis et~al.(2021)Dimitriadis, Gneiting, and Jordan]{dimitriadis2021stable}
Timo Dimitriadis, Tilmann Gneiting, and Alexander~I. Jordan.
\newblock Stable reliability diagrams for probabilistic classifiers.
\newblock \emph{Proceedings of the National Academy of Sciences}, 118\penalty0 (8):\penalty0 e2016191118, 2021.
\newblock \doi{10.1073/pnas.2016191118}.

\bibitem[Du et~al.(2025)Du, Xu, Zhu, Wang, and Mao]{deepresearchbench2025}
Mingxuan Du, Benfeng Xu, Chiwei Zhu, Xiaorui Wang, and Zhendong Mao.
\newblock {DeepResearch Bench}: A comprehensive benchmark for deep research agents, 2025.
\newblock \url{https://arxiv.org/abs/2506.11763}.

\bibitem[Grattafiori et~al.(2024)Grattafiori, Dubey, Jauhri, et~al.]{grattafiori2024llama3}
Aaron Grattafiori, Abhimanyu Dubey, Abhinav Jauhri, et~al.
\newblock The {Llama 3} herd of models, 2024.
\newblock \url{https://arxiv.org/abs/2407.21783}.

\bibitem[Hill and Stein(2025)]{hill2025competition}
Ryan Hill and Carolyn Stein.
\newblock Race to the bottom: Competition and quality in science.
\newblock \emph{The Quarterly Journal of Economics}, 140\penalty0 (2):\penalty0 1111--1185, 2025.
\newblock \doi{10.1093/qje/qjaf010}.

\bibitem[Hu et~al.(2026)Hu, Zhao, Zhang, Cohan, and Zhao]{sage2026}
Tiansheng Hu, Yilun Zhao, Canyu Zhang, Arman Cohan, and Chen Zhao.
\newblock {SAGE}: Benchmarking and improving retrieval for deep research agents, 2026.
\newblock \url{https://arxiv.org/abs/2602.05975}.

\bibitem[Hyndman and Athanasopoulos(2021)]{hyndman2021forecasting}
Rob~J. Hyndman and George Athanasopoulos.
\newblock \emph{Forecasting: Principles and Practice}.
\newblock OTexts, Melbourne, Australia, 3 edition, 2021.
\newblock \url{https://otexts.com/fpp3/}.

\bibitem[Li et~al.(2025)Li, Xu, Guo, Zhao, Li, Yuan, Zhang, Jiang, Xin, Dang, Rong, Zhao, Feng, and Bing]{coi2025}
Long Li, Weiwen Xu, Jiayan Guo, Ruochen Zhao, Xingxuan Li, Yuqian Yuan, Boqiang Zhang, Yuming Jiang, Yifei Xin, Ronghao Dang, Yu~Rong, Deli Zhao, Tian Feng, and Lidong Bing.
\newblock Chain of ideas: Revolutionizing research via novel idea development with {LLM} agents.
\newblock In \emph{Findings of the Association for Computational Linguistics: EMNLP 2025}, pages 8971--9004, Suzhou, China, 2025. Association for Computational Linguistics.
\newblock \doi{10.18653/v1/2025.findings-emnlp.477}.

\bibitem[Luo et~al.(2025)Luo, Rechardt, Sun, et~al.]{luo2025brainbench}
Xiaoliang Luo, Akilles Rechardt, Guangzhi Sun, et~al.
\newblock Large language models surpass human experts in predicting neuroscience results.
\newblock \emph{Nature Human Behaviour}, 9\penalty0 (2):\penalty0 305--315, 2025.
\newblock \doi{10.1038/s41562-024-02046-9}.

\bibitem[Marchionini(2006)]{marchionini2006exploratory}
Gary Marchionini.
\newblock Exploratory search: from finding to understanding.
\newblock \emph{Communications of the ACM}, 49\penalty0 (4):\penalty0 41--46, 2006.
\newblock \doi{10.1145/1121949.1121979}.

\bibitem[{Meta}(2024{\natexlab{a}})]{llama31modelcard2024}
{Meta}.
\newblock {Llama 3.1 8B Instruct} model card, 2024{\natexlab{a}}.
\newblock \url{https://huggingface.co/meta-llama/Llama-3.1-8B-Instruct}.
\newblock Accessed 2026-10-15; documented pretraining cutoff: December 2023.

\bibitem[{Meta}(2024{\natexlab{b}})]{llama32modelcard2024}
{Meta}.
\newblock {Llama 3.2 3B Instruct} model card, 2024{\natexlab{b}}.
\newblock \url{https://huggingface.co/meta-llama/Llama-3.2-3B-Instruct}.
\newblock Accessed 2026-10-15; documented pretraining cutoff: December 2023.

\bibitem[Murphy(1973)]{murphy1973partition}
Allan~H. Murphy.
\newblock A new vector partition of the probability score.
\newblock \emph{Journal of Applied Meteorology}, 12\penalty0 (4):\penalty0 595--600, 1973.
\newblock \doi{10.1175/1520-0450(1973)012<0595:ANVPOT>2.0.CO;2}.

\bibitem[{OpenAI}(2025{\natexlab{a}})]{openaioss2025}
{OpenAI}.
\newblock Introducing gpt-oss, August 2025{\natexlab{a}}.
\newblock \url{https://openai.com/index/introducing-gpt-oss/}.
\newblock Published 5 August 2025; accessed 25 October 2026.

\bibitem[{OpenAI}(2025{\natexlab{b}})]{openaiossmodelcard2025}
{OpenAI}.
\newblock {gpt-oss-120b} \& {gpt-oss-20b} model card, 2025{\natexlab{b}}.
\newblock \url{https://arxiv.org/abs/2508.10925}.

\bibitem[{OpenAI}(2026{\natexlab{a}})]{openaiapichangelog2026}
{OpenAI}.
\newblock {OpenAI API} changelog, 2026{\natexlab{a}}.
\newblock \url{https://developers.openai.com/api/docs/changelog}.
\newblock GPT-5.6 family released 9 July 2026; GPT-6 Sol and Luna released 22 September 2026; accessed 26 October 2026.

\bibitem[{OpenAI}(2026{\natexlab{b}})]{openaigpt56systemcard2026}
{OpenAI}.
\newblock {GPT-5.6} system card.
\newblock Technical report, OpenAI, July 2026{\natexlab{b}}.
\newblock \url{https://deploymentsafety.openai.com/gpt-5-6}.
\newblock Accessed 26 October 2026.

\bibitem[{OpenAI}(2026{\natexlab{c}})]{openaigpt56terra2026}
{OpenAI}.
\newblock {GPT-5.6 Terra} model documentation, 2026{\natexlab{c}}.
\newblock \url{https://developers.openai.com/api/docs/models/gpt-5.6-terra}.
\newblock Accessed 2026-10-26; documented knowledge cutoff: 2026-02-16.

\bibitem[{OpenAI}(2026{\natexlab{d}})]{openaigpt6luna2026}
{OpenAI}.
\newblock {GPT-6 Luna} model documentation, 2026{\natexlab{d}}.
\newblock \url{https://developers.openai.com/api/docs/models/gpt-6-luna}.
\newblock Accessed 2026-10-26; documented knowledge cutoff: 2026-05-18.

\bibitem[{OpenAI}(2026{\natexlab{e}})]{openaigpt6sol2026}
{OpenAI}.
\newblock {GPT-6 Sol} model documentation, 2026{\natexlab{e}}.
\newblock \url{https://developers.openai.com/api/docs/models/gpt-6-sol}.
\newblock Accessed 2026-10-26; documented knowledge cutoff: 2026-04-20.

\bibitem[{OpenAI}(2026{\natexlab{f}})]{openaigpt6systemcard2026}
{OpenAI}.
\newblock {GPT-6 Astra} system card.
\newblock Technical report, OpenAI, September 2026{\natexlab{f}}.
\newblock \url{https://deploymentsafety.openai.com/gpt-6-astra/sec:appendix-sol-luna}.
\newblock Appendix: GPT-6 Sol, GPT-6 Luna (22 September 2026). Accessed 26 October 2026.

\bibitem[{Qwen Team}(2025{\natexlab{a}})]{qwen3release2025}
{Qwen Team}.
\newblock {Qwen3}: Think deeper, act faster, April 2025{\natexlab{a}}.
\newblock \url{https://qwenlm.github.io/blog/qwen3/}.
\newblock Published 29 April 2025; accessed 24 October 2026.

\bibitem[{Qwen Team}(2025{\natexlab{b}})]{qwen3technicalreport2025}
{Qwen Team}.
\newblock {Qwen3} technical report, 2025{\natexlab{b}}.
\newblock \url{https://arxiv.org/abs/2505.09388}.

\bibitem[{Qwen Team}(2026{\natexlab{a}})]{qwen35modelcard2026}
{Qwen Team}.
\newblock {Qwen3.5-4B} model card, 2026{\natexlab{a}}.
\newblock \url{https://huggingface.co/Qwen/Qwen3.5-4B}.
\newblock Accessed 2026-10-15.

\bibitem[{Qwen Team}(2026{\natexlab{b}})]{qwen38modelcard2026}
{Qwen Team}.
\newblock {Qwen3.8-27B} model card, 2026{\natexlab{b}}.
\newblock \url{https://huggingface.co/Qwen/Qwen3.8-27B}.
\newblock Accessed 2026-10-15.

\bibitem[{Qwen Team}(2026{\natexlab{c}})]{qwenreleases2026}
{Qwen Team}.
\newblock {Qwen} open model release announcements, 2026{\natexlab{c}}.
\newblock \url{https://github.com/QwenLM/Qwen3.8\#news}.
\newblock Accessed 2026-10-15; Qwen3.5-4B released 2026-03-02 and Qwen3.8-27B released 2026-08-14.

\bibitem[Rafols et~al.(2010)Rafols, Porter, and Leydesdorff]{rafols2010overlay}
Ismael Rafols, Alan~L. Porter, and Loet Leydesdorff.
\newblock Science overlay maps: A new tool for research policy and library management.
\newblock \emph{Journal of the American Society for Information Science and Technology}, 61\penalty0 (9):\penalty0 1871--1887, 2010.
\newblock \doi{10.1002/asi.21368}.

\bibitem[Rahkovsky et~al.(2021)Rahkovsky, Toney, Boyack, Klavans, and Murdick]{rahkovsky2021portfolios}
Ilya Rahkovsky, Autumn Toney, Kevin~W. Boyack, Richard Klavans, and Dewey~A. Murdick.
\newblock {AI} research funding portfolios and extreme growth.
\newblock \emph{Frontiers in Research Metrics and Analytics}, 6:\penalty0 630124, 2021.
\newblock \doi{10.3389/frma.2021.630124}.

\bibitem[Skarlinski et~al.(2024)Skarlinski, Cox, Laurent, Braza, Hinks, Hammerling, Ponnapati, Rodriques, and White]{paperqa2024}
Michael~D. Skarlinski, Sam Cox, Jon~M. Laurent, James~D. Braza, Michaela Hinks, Michael~J. Hammerling, Manvitha Ponnapati, Samuel~G. Rodriques, and Andrew~D. White.
\newblock Language agents achieve superhuman synthesis of scientific knowledge, 2024.
\newblock \url{https://arxiv.org/abs/2409.13740}.

\bibitem[Tian et~al.(2026)Tian, Yin, Su, Chao, Gu, Kong, Xia, and Liu]{foresci2026}
Qiuyu Tian, Haojie Yin, Hang Su, Jianghan Chao, Xiaowen Gu, Youyong Kong, Yingce Xia, and Zequn Liu.
\newblock {ForeSci}: Evaluating {LLM} agents for forward-looking {AI} research judgment, 2026.
\newblock \url{https://arxiv.org/abs/2606.00644}.

\bibitem[Tong et~al.(2026)Tong, Li, Li, Yang, Mou, Ma, Chen, Liu, Cheng, Zhang, Chen, Ge, Guo, Ying, Sun, Zheng, Xi, Chen, Zhao, Ding, Huang, Jiang, and Qiu]{tong2026ailearnscientifictaste}
Jingqi Tong, Mingzhe Li, Hangcheng Li, Yongzhuo Yang, Yurong Mou, Weijie Ma, Hongji Chen, Xiaoran Liu, Qinyuan Cheng, Ming Zhang, Qiguang Chen, Weifeng Ge, Qipeng Guo, Tianlei Ying, Tianxiang Sun, Yining Zheng, Zhiheng Xi, Xinchi Chen, Jun Zhao, Ning Ding, Xuanjing Huang, Yu-Gang Jiang, and Xipeng Qiu.
\newblock {AI} can learn scientific taste.
\newblock \emph{arXiv preprint arXiv:2603.14473}, 2026.
\newblock \url{https://arxiv.org/abs/2603.14473}.

\bibitem[Wu et~al.(2026)Wu, Lu, Chen, Bragg, Yamada, Clark, Clifton, Torr, Zou, and Yu]{wu2026cusp}
Sean Wu, Pan Lu, Yupeng Chen, Jonathan Bragg, Yutaro Yamada, Peter Clark, David Clifton, Philip Torr, James Zou, and Junchi Yu.
\newblock Scientific reasoning does not reliably translate into scientific forecasting in frontier {AI}, 2026.
\newblock \url{https://arxiv.org/abs/2605.22681}.

\bibitem[Xu et~al.(2025)Xu, Lu, Ye, Hu, and Liu]{researcherbench2025}
Tianze Xu, Pengrui Lu, Lyumanshan Ye, Xiangkun Hu, and Pengfei Liu.
\newblock {ResearcherBench}: Evaluating deep {AI} research systems on the frontiers of scientific inquiry, 2025.
\newblock \url{https://arxiv.org/abs/2507.16280}.

\bibitem[Yue et~al.(2025)Yue, Wang, Chen, Huang, and Wei]{yue2025synergistic}
Shengbin Yue, Siyuan Wang, Wei Chen, Xuanjing Huang, and Zhongyu Wei.
\newblock Synergistic multi-agent framework with trajectory learning for knowledge-intensive tasks.
\newblock \emph{Proceedings of the AAAI Conference on Artificial Intelligence}, 39\penalty0 (24):\penalty0 25796--25804, 2025.
\newblock \doi{10.1609/aaai.v39i24.34772}.

\bibitem[Yue et~al.(2026)Yue, Wang, Wang, Chen, Chen, and Wei]{yue2026paraagent}
Shengbin Yue, Hongru Wang, Siyuan Wang, Xiaoxin Chen, Wei Chen, and Zhongyu Wei.
\newblock {ParaAgent}: Reinforcing parallel acting in open-world tool environments.
\newblock \emph{arXiv preprint arXiv:2609.33618}, 2026.
\newblock \url{https://arxiv.org/abs/2609.33618}.

\end{thebibliography}
\clearpage
\beginappendix
\section{Construction and Reference Protocol}
\label{app:benchmark-details}

\subsection{Corpus, Fields and Direction Cards}
\label{app:construction}

\paragraph{Paper records and field seeds.}
The source corpus comprises 356,357 arXiv papers first submitted between June 2022 and June 2026, tagged with at least one of cs.CL, cs.LG, cs.AI, or cs.CV. Qwen3.8-27B extracts a field, research problem, technical focus, setting, and direction from each complete title and abstract, without input truncation. Field construction uses records first submitted before January 2024. GPT-6-sol normalises recurring field names; a minimum of 50 construction papers supplies the initial field seeds. Candidate retrieval over paper text and extracted labels is followed by Qwen-based paper--field classification, rather than asking the classifier to choose from the entire field inventory. After merging three pairs of name variants, the inventory contains 271 fields. A paper may belong to more than one field.

\paragraph{Fine-grained direction cards.}
Within each field, embedding-based neighbourhoods organise the historical construction papers into small evidence groups. GPT-6-sol identifies related research problems and consolidates them into direction cards specifying names, scope, inclusion and exclusion criteria, and supporting papers. Construction-stage assignments expose coverage gaps, inconsistent boundaries, and redundant parent--child definitions for revision. These fine-grained cards and their paper links constitute the environment's retained research structure.

\paragraph{Consolidation.}
Shared examples from image restoration, point-cloud processing, and speech recognition guide consolidation into broader research-problem families. Consolidation allows at most 15 benchmark directions per field, with shared examples guiding granularity and no prescribed exact count or balanced class size. Each fine-grained direction maps to exactly one parent, whose scope is the union of its components. A paper need not address every component to belong to the parent. Fine-grained definitions and mappings remain available independently of the benchmark's candidate selection.

\paragraph{Paper memberships.}
Qwen3.8-27B applies the frozen fine-grained cards to complete titles and abstracts without paper dates or expected activity shares. Classification distinguishes confirmed memberships, uncertain memberships, and papers without a supported direction. Multiple confirmed memberships are permitted, including when the same central contribution satisfies more than one direction. Benchmark memberships are obtained by taking the union of confirmed fine-grained memberships and deduplicating parent IDs. Fractional weights are then recomputed at the benchmark level; summing already-normalised child shares would produce a different measure. Papers without a confirmed direction remain in the coverage audit rather than being forced into an \textsc{other} answer category.

\begin{figure*}[p]
\centering
\includegraphics[width=\textwidth,height=0.82\textheight,keepaspectratio]{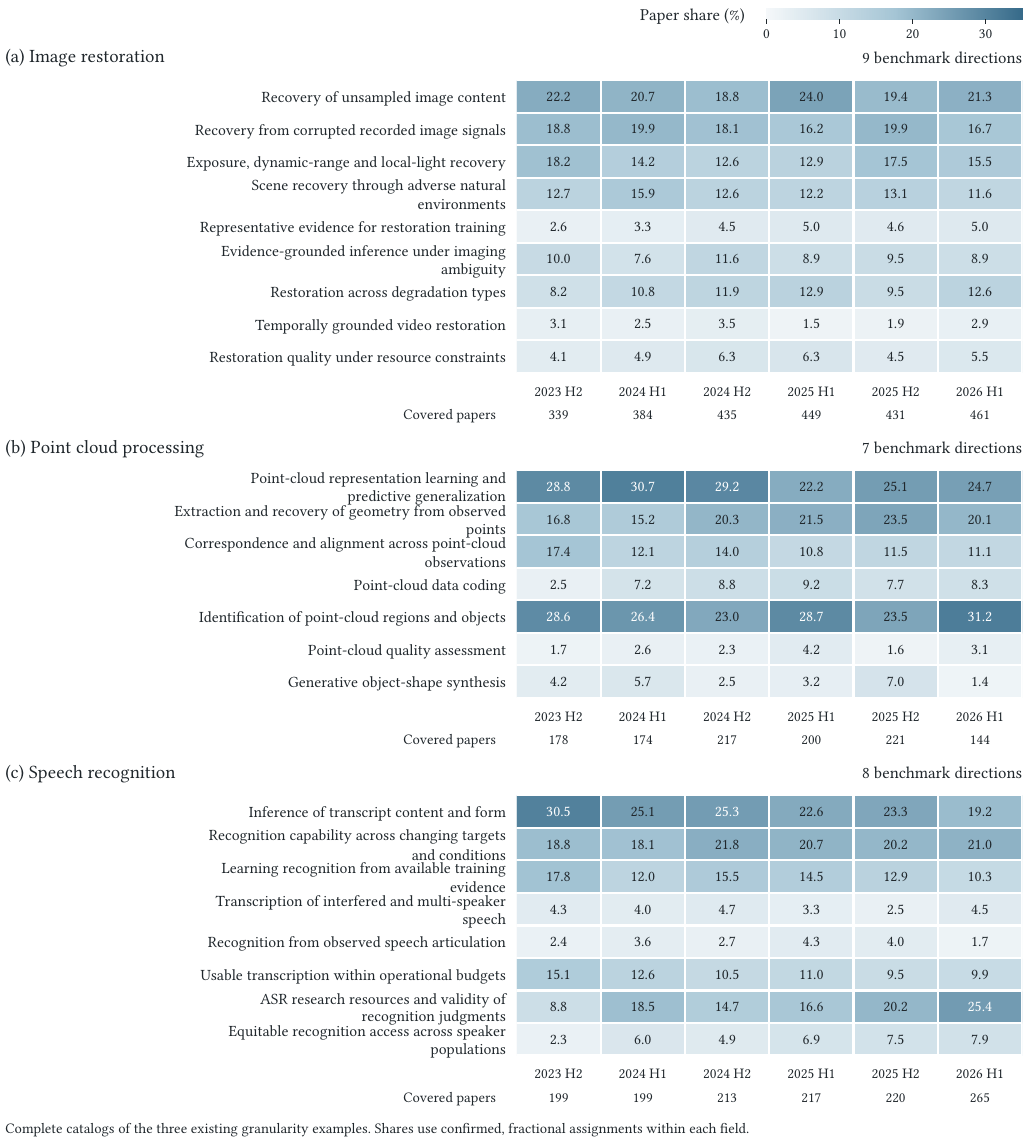}
\caption{Complete benchmark direction catalogues for the three consolidation examples: image restoration, point cloud processing and speech recognition. Cells show field-wide paper shares (\%) in six half-years, from 2023 H2 through 2026 H1, using a common colour scale and confirmed fractional memberships. Each covered paper contributes unit mass divided across its confirmed benchmark directions. The covered-paper counts beneath each panel give the corresponding denominators.}
\Description{Three heatmaps list all nine image-restoration directions, seven point-cloud-processing directions and eight speech-recognition directions. Six columns show half-year paper shares from 2023 H2 to 2026 H1, with covered-paper counts beneath each panel and a shared colour scale.}
\label{fig:direction-examples}
\end{figure*}

Figure~\ref{fig:direction-examples} shows the resulting catalogues for the three shared consolidation examples, with nine, seven and eight benchmark directions, respectively. These fields were used to guide granularity before evaluation, rather than selected for model performance or large observed changes.

\subsection{Aggregation and Reference Answers}
\label{app:references}
Let $\mathcal P_f^+(W)$ contain papers in field $f$ first submitted in window $W$ with at least one confirmed benchmark direction, and let $A_f(p)$ contain paper $p$'s distinct confirmed directions. Direction $d$ has share $s_{f,d}(W)$, given by
\begin{equation}
 s_{f,d}(W)=\frac{1}{|\mathcal P_f^+(W)|}
 \sum_{p\in\mathcal P_f^+(W)}\frac{\mathbf 1\{d\in A_f(p)\}}{|A_f(p)|}.
 \label{eq:activity-share}
\end{equation}
The indicator equals one for membership in $d$ and zero otherwise; vertical bars denote set size. Candidate selection leaves these field-wide weights and denominator unchanged. Trend references within $10^{-12}$ of zero and empty covered windows are unscored. Comparison accepts all candidates tied for the largest signed change within this tolerance. Small nonzero changes remain in evaluation.

\subsection{Sampling and Temporal Alignment}
\label{app:rolling}
A fixed-seed hash ordering selects up to three directions per field, without reading shares, reference answers or model scores. Each supplies a Trend question; their set supplies one Comparison question. Candidate identity and order stay fixed across dates and targets. The 54 Core fields are hash-sampled within directory-size strata 2--5, 6--10 and 11--15, with quotas 14, 26 and 14. Two single-direction fields are excluded because their covered share is always one. Each episode starts independently. Main results pair Forecast at cut-off $T$ with State at $T+6$ months; the extra January 2024 State round completes the five same-cut-off diagnostic pairs and is excluded from main scores.

The five paired target changes are listed in Table~\ref{tab:rolling-windows}. The same-event comparison shifts the State cut-off forward by six months; the same-cut-off comparison retains one origin for both temporal targets. They reuse the selected directions but answer different analytical questions.

\begin{table*}[t]
\centering\small
\caption{Rolling windows used for the main same-event comparison. Each row pairs two questions about the same change in paper share. Search accepts papers strictly before the listed cut-off.}
\label{tab:rolling-windows}
\begin{tabular}{@{}llll@{}}
\toprule
Earlier reference window & Later reference window & Forecast cut-off & State cut-off \\
\midrule
July--December 2023 & January--June 2024 & 2024-01-01 & 2024-07-01 \\
January--June 2024 & July--December 2024 & 2024-07-01 & 2025-01-01 \\
July--December 2024 & January--June 2025 & 2025-01-01 & 2025-07-01 \\
January--June 2025 & July--December 2025 & 2025-07-01 & 2026-01-01 \\
July--December 2025 & January--June 2026 & 2026-01-01 & 2026-07-01 \\
\bottomrule
\end{tabular}
\end{table*}

\subsection{Examples and Core Instructions}
\label{app:task-examples}
Table~\ref{tab:prompt-examples} reproduces frozen question texts and system-instruction excerpts for 3D human reconstruction at 1 July 2024. Each request includes the field, cut-off, and selected directions' names and scopes. The Comparison scopes are reproduced below; the Trend question receives only its selected direction. IDs are local to each question.

\noindent\textbf{D1: Creation of new 3D human identities.} Create 3D human-body or head identities and assets rather than recover observed people. This includes learning reusable population-level generative distributions and constructing specific renderable characters from text, pose, or other creation guidance.

\noindent\textbf{D2: Reconstruction of controllable clothing.} Obtain clothing that can be treated as an independent object, and its relationship to the body, from observations of a person. This includes separating replaceable body and garment layers and recovering garment surfaces and deformation behavior as pose or time changes.

\noindent\textbf{D3: Recovery of observed human form and motion.} Recover the 3D form, pose, or motion of a specific person from visual observations, including parametric bodies and meshes, static faces and heads, complete high-detail clothed surfaces, and reconstruction constrained by contact with other people, objects, or scenes.

\begin{table*}[t]
\caption{Frozen question texts and selected Agent instructions for one field and cut-off. The two Comparison questions share the displayed candidates; direction scopes appear in the accompanying text; remaining system clauses are omitted.}
\label{tab:prompt-examples}
\centering\small
\begin{tabularx}{\textwidth}{@{}XX@{}}
\toprule
\textbf{Shared system instructions (excerpts).}\par
Research attention is measured by a direction's share of arXiv papers within its field. Papers belonging to multiple directions are counted fractionally. The target is paper share, not research quality or scientific value. Use Search to investigate, then submit the probability object through Finish. Make at least one successful Search call before Finish. You may call Search at most 20 times; each returns up to 8 matching papers with full titles, abstracts and total matches.
\par\medskip\textbf{Trend output (excerpt).}\par
Return \{"p\_gain": p\}, where p is your probability that the direction's field-wide paper share increased from the earlier period to the later period. Use a number from 0 to 1; the probability of a decrease is 1-p.
\par\medskip\textbf{Comparison output (excerpt).}\par
Return \{"probabilities": \{direction\_id: probability, ...\}\} for every listed candidate. Each probability is the chance that this candidate has the largest signed change in field-wide paper share: later period minus earlier period. If every candidate declines, the winner has the smallest decline. Compare percentage-point changes, not percentage growth rates or final share levels. Use numbers from 0 to 1 summing to 1. These are probabilities of winning, not paper shares. &
\textbf{Question texts.}\par
\textit{State / Trend.} During January--June 2024, was research on "Recovery of observed human form and motion" getting more or less attention within 3d human reconstruction than in July--December 2023?\par\smallskip
\textit{Forecast / Trend.} Based on the literature available before 2024-07-01, do you expect research on "Recovery of observed human form and motion" to receive more or less attention within 3d human reconstruction during July--December 2024 than in January--June 2024?\par\smallskip
\textit{State / Comparison.} Compared with July--December 2023, which of these research directions gained the most attention within 3d human reconstruction during January--June 2024?\par\smallskip
\textit{Forecast / Comparison.} Based on the literature available before 2024-07-01, which of these research directions do you expect to gain the most attention within 3d human reconstruction during July--December 2024, compared with January--June 2024?\par\smallskip
\textbf{Comparison candidates.} D1: Creation of new 3D human identities\newline D2: Reconstruction of controllable clothing\newline D3: Recovery of observed human form and motion \\
\bottomrule
\end{tabularx}
\end{table*}

\paragraph{Search interface.}
\texttt{Search} requires a plain-text \texttt{query} and \texttt{sort}, either \texttt{relevance} or \texttt{date}. Optional \texttt{date\_from} is inclusive and \texttt{date\_to} exclusive, in YYYY-MM-DD format. The server enforces the question's field and pre-cut-off access even if a query requests a wider interval. Results contain paper IDs, dates, full titles and abstracts, and the total matching-record count; paper--direction memberships are withheld.

\paragraph{Outputs.}
\label{app:answers}
Trend returns $p_{\mathrm{gain}}$, the probability of gain, and selects gain when $p_{\mathrm{gain}}\geq0.5$. Comparison returns probabilities for all candidates and selects their maximum, breaking probability ties lexically by direction ID. These probabilities describe answer events, not paper shares.

\section{Reference Validity and Robustness}
\label{app:reference-robustness}

\paragraph{Human audit protocol.}
The four fields in Table~\ref{tab:human-model-agreement} were selected for annotator familiarity. Within each field, we sampled 25 papers from July--December 2025 and January--June 2026 using fixed-seed hash ranking within half-year and machine-membership strata (single, multiple or no confirmed directions); unfilled quotas were reallocated to remaining strata. The annotation interface displayed complete titles, abstracts and frozen direction definitions, with model assignments and benchmark answers hidden. The two researchers independently selected all directions supported by a paper's central research contribution; incidental method use or keyword overlap did not qualify.

\paragraph{Membership and counting sensitivity.}
Adding a second membership halves a paper's original fractional weight. Giving each confirmed membership unit weight and normalising over all memberships changes 9.1\% of Trend answers and 13.0\% of Comparison acceptable-answer sets on jointly scorable Core events (field-macro); three Trend events become unscorable ties. Re-scoring all three Closed and Agent runs for Sol, Qwen3.8, DeepSeek and gpt-oss preserves the larger State search gain for every model and task. Counts use each realised event once across State and Forecast.

\paragraph{Coverage and margins.}
\label{app:boundary-coverage}
Coverage is the fraction of field papers with a confirmed direction. Its mean across 269 fields is 92.1--93.1\% over the six half-years, while 8--19 fields fall below 80\% in each window. Including uncovered papers in the denominator changes about 3\% of answers (Table~\ref{tab:robustness-summary}); every Core accuracy change stays below one percentage point across the four models.

\begin{table}[t]
\centering\small
\caption{Sensitivity checks on unchanged outputs from three runs of four models. Reference changes use unique events.}
\label{tab:robustness-summary}
\begin{tabularx}{\columnwidth}{@{}lX@{}}
\toprule
Check & Result \\
\midrule
Full membership weights & Core reference changes: 9.1\% Trend, 13.0\% Comparison (field-macro). \\
All-paper denominator & Core changes: 2.84\% / 3.33\%; Extended: 3.23\% / 3.27\% (Trend / Comparison, event-weighted). \\
Minimum margin & State search gains exceed Forecast gains at 0, 0.5, 1, 2 and 5 pp on both tasks. \\
Coverage $\geq90\%$ & The contrast persists on 570 Trend and 190 Comparison events per target. \\
\bottomrule
\end{tabularx}
\end{table}

The margin is absolute share change for Trend and the winning change minus the runner-up for Comparison. Respectively, 30.0\% and 25.9\% of Core events have margins below one percentage point. Coverage filters require the threshold in both half-years. These filters affect diagnostic subsets only. Date-alignment conclusions also persist with a one-point margin and the all-paper denominator: State retains positive alignment advantages on both tasks for all four models; Forecast advantages are near zero or negative.

\paragraph{Denominator sensitivity by margin.}
The alternative denominator retains each direction's confirmed fractional mass, but divides it by all field papers in the window. Uncovered papers retain unassigned mass; the named directions are not renormalised. Table~\ref{tab:denominator-sensitivity} separates all events from those whose original reference margin is at least one percentage point. Across the eight Core accuracy cells per model, the maximum absolute change is 0.74, 0.88, 0.99 and 0.99 points for Sol, Qwen3.8-27B, DeepSeek and gpt-oss, respectively.
\begin{table*}[t]
\caption{Reference-answer sensitivity to including uncovered papers in the denominator. Changed denotes a different acceptable-answer set, including transitions to or from an exact tie. Percentages weight events equally. Margins and diagnostic subsets are fixed using the original references; pp denotes percentage points.}
\label{tab:denominator-sensitivity}
\newcommand{\rapDenominatorSensitivityRows}{%
Core & Trend & 810 & 2.84 & 567 & 0.53 \\
Core & Comparison & 270 & 3.33 & 200 & 0.50 \\
\addlinespace[2pt]
Extended & Trend & 4,025 & 3.23 & 2,822 & 0.99 \\
Extended & Comparison & 1,345 & 3.27 & 997 & 1.10 \\
}

\centering\small
\setlength{\tabcolsep}{10pt}
\begin{tabular}{@{}llrrrr@{}}
\toprule
& & \multicolumn{2}{c}{\textbf{All events}} & \multicolumn{2}{c}{\textbf{Reference margin $\geq 1$ pp}} \\
\cmidrule(lr){3-4}\cmidrule(l){5-6}
\textbf{Set} & \textbf{Task} & \textbf{Events} & \textbf{Changed (\%)} & \textbf{Events} & \textbf{Changed (\%)} \\
\midrule
\rapDenominatorSensitivityRows
\bottomrule
\end{tabular}
\end{table*}

\section{Evaluation Details}
\label{app:evaluation-details}

\paragraph{Execution and fixed search.}
The context and per-call output limits are 65,536 and 16,384 tokens. Oldest tool bodies are omitted first; past reasoning is omitted except for DeepSeek's required native reasoning fields. Fixed search prepares twenty queries from field and direction names before inference, distributed across two completed half-years for State and three for Forecast. Each returns up to eight full titles and abstracts under the same index and cut-off as Agent, followed by one Sol judgment. The context policy is shared; evaluator memberships and shares are withheld.

\paragraph{Scoring.}
For question $i$, let $\ell_i$ denote its Brier loss, with the superscript specifying the task:
\begin{equation}
 \ell_i^{\mathrm{Trend}}=(p_i-y_i)^2,\qquad
 \ell_i^{\mathrm{Comparison}}=\tfrac12\sum_{c\in C_i}(p_{ic}-\mathbf1\{y_i=c\})^2.
 \label{eq:brier-loss}
\end{equation}
For Trend, $p_i$ is the probability of gain and $y_i$ its reference (1 for gain, 0 for loss). For Comparison, $C_i$ is the candidate set, $c$ indexes a candidate, $p_{ic}$ is its predicted probability, and $y_i$ is the reference winner. The indicator $\mathbf1\{y_i=c\}$ is one when $c$ wins and zero otherwise. Both losses lie in $[0,1]$; lower is better. Invalid outputs count as incorrect for accuracy and are excluded from Brier; tied Comparison references are also excluded from Brier. Scores average within fields, then equally across fields.

\paragraph{Output validation and repeats.}
\label{app:execution-analysis}
Main Agent answers require a native Finish after successful Search. A shared validator checks keys, finite probabilities and bounds, accepts numeric strings, and normalises sums within 0.01 of one. DeepSeek's Closed Comparison validity is 30.0\% for State and 31.0\% for Forecast, versus 99.6\% and 99.1\% in Agent. Main scores average three complete runs: Agent regenerates entire trajectories, Closed regenerates answers, and fixed search reuses its evidence. Across model--task cells, accuracy standard deviations range from 0.07--3.76 points for Closed and 0.21--4.90 for Agent; Sol fixed search ranges from 0.37--1.07. Final-judgment repeats reuse a fixed search history.

Tables~\ref{tab:main-brier} and~\ref{tab:completed-repeats} provide probability losses and the standard deviation of accuracy across the three complete runs. Brier loss conditions on valid outputs with unique references, so its eligible questions can differ across models and resource settings. Agent repeats regenerate complete investigations; fixed-search repeats retain the same retrieved evidence.

\begin{table}[t]
\caption{Core Brier loss averaged over the complete runs in Table~\ref{tab:core-main}, using valid outputs with unique references; lower is better. Validity differs across models.}
\label{tab:main-brier}
\newcommand{\rapCurrentBrierRows}{%
\multicolumn{5}{@{}l}{\textit{Closed}} \\
Llama-3.2-3B-Instruct & 0.289 & 0.381 & 0.284 & 0.370 \\
Llama-3.1-8B-Instruct & 0.288 & 0.430 & 0.286 & 0.379 \\
Qwen3-4B & 0.267 & 0.397 & 0.276 & 0.408 \\
Qwen3-8B & 0.263 & 0.378 & 0.273 & 0.387 \\
Qwen3.5-4B & 0.259 & 0.364 & 0.267 & 0.361 \\
Qwen3.8-27B & 0.260 & 0.384 & 0.258 & 0.378 \\
gpt-oss-120b & 0.268 & 0.388 & 0.273 & 0.375 \\
DeepSeek-V4.1-Flash & 0.268 & 0.356 & 0.266 & 0.358 \\
GPT-5.6-terra & 0.252 & 0.368 & 0.254 & 0.366 \\
GPT-6-luna & 0.258 & 0.381 & 0.259 & 0.380 \\
GPT-6-sol & 0.256 & 0.361 & 0.255 & 0.367 \\
\multicolumn{5}{@{}l}{\textit{Agent}} \\
Llama-3.2-3B-Instruct & 0.460 & 0.384 & 0.458 & 0.361 \\
Llama-3.1-8B-Instruct & 0.412 & 0.423 & 0.429 & 0.369 \\
Qwen3-4B & 0.393 & 0.466 & 0.305 & 0.383 \\
Qwen3-8B & 0.280 & 0.437 & 0.291 & 0.395 \\
Qwen3.5-4B & 0.331 & 0.364 & 0.304 & 0.362 \\
Qwen3.8-27B & 0.223 & 0.309 & 0.272 & 0.384 \\
gpt-oss-120b & 0.260 & 0.364 & 0.295 & 0.378 \\
DeepSeek-V4.1-Flash & 0.220 & 0.306 & 0.279 & 0.398 \\
GPT-5.6-terra & 0.233 & 0.347 & 0.268 & 0.373 \\
GPT-6-luna & 0.265 & 0.371 & 0.286 & 0.389 \\
GPT-6-sol & 0.190 & 0.308 & 0.262 & 0.394 \\
\multicolumn{5}{@{}l}{\textit{Fixed search}} \\
GPT-6-sol & 0.205 & 0.306 & 0.267 & 0.391 \\
}

\centering\small
\setlength{\tabcolsep}{3pt}
\begin{tabular}{@{}lrrrr@{}}
\toprule
& \multicolumn{2}{c}{\textbf{State}} & \multicolumn{2}{c}{\textbf{Forecast}} \\
\cmidrule(lr){2-3}\cmidrule(lr){4-5}
\textbf{Model} & \textbf{Trend} & \textbf{Comparison} & \textbf{Trend} & \textbf{Comparison} \\
\midrule
\rapCurrentBrierRows
\bottomrule
\end{tabular}
\end{table}

\begin{table*}[t]
\caption{Sample standard deviation of Core field-macro accuracy across independent runs, in percentage points. Each run counts invalid answers as wrong; every entry uses three complete runs.}
\label{tab:completed-repeats}
\newcommand{\rapRunDeviationRows}{%
\multicolumn{6}{@{}l}{\textit{Closed}} \\
Llama-3.2-3B-Instruct & 3 & 0.57 & 2.60 & 2.14 & 1.67 \\
Llama-3.1-8B-Instruct & 3 & 0.91 & 2.78 & 1.43 & 3.76 \\
Qwen3-4B & 3 & 0.31 & 1.13 & 0.94 & 1.67 \\
Qwen3-8B & 3 & 0.71 & 1.70 & 0.07 & 1.13 \\
Qwen3.5-4B & 3 & 0.70 & 2.11 & 2.66 & 0.77 \\
Qwen3.8-27B & 3 & 1.31 & 1.07 & 1.62 & 2.43 \\
gpt-oss-120b & 3 & 0.93 & 2.35 & 0.19 & 1.54 \\
DeepSeek-V4.1-Flash & 3 & 0.78 & 1.75 & 0.56 & 1.13 \\
GPT-5.6-terra & 3 & 2.11 & 1.50 & 1.12 & 1.92 \\
GPT-6-luna & 3 & 0.78 & 1.90 & 0.29 & 1.61 \\
GPT-6-sol & 3 & 1.35 & 0.98 & 0.47 & 1.19 \\
\multicolumn{6}{@{}l}{\textit{Agent}} \\
Llama-3.2-3B-Instruct & 3 & 0.96 & 3.53 & 2.40 & 1.96 \\
Llama-3.1-8B-Instruct & 3 & 0.45 & 1.13 & 0.93 & 4.90 \\
Qwen3-4B & 3 & 0.31 & 0.21 & 0.96 & 2.80 \\
Qwen3-8B & 3 & 1.51 & 0.77 & 2.28 & 1.34 \\
Qwen3.5-4B & 3 & 0.33 & 1.86 & 0.31 & 2.47 \\
Qwen3.8-27B & 3 & 1.19 & 1.70 & 0.74 & 2.60 \\
gpt-oss-120b & 3 & 0.31 & 1.19 & 1.30 & 3.87 \\
DeepSeek-V4.1-Flash & 3 & 0.50 & 3.85 & 0.99 & 0.77 \\
GPT-5.6-terra & 3 & 1.22 & 1.85 & 0.69 & 1.54 \\
GPT-6-luna & 3 & 0.57 & 2.97 & 0.64 & 1.30 \\
GPT-6-sol & 3 & 0.75 & 0.64 & 0.40 & 1.13 \\
\multicolumn{6}{@{}l}{\textit{Fixed search}} \\
GPT-6-sol & 3 & 0.38 & 1.07 & 0.71 & 0.37 \\
}

\centering\small
\setlength{\tabcolsep}{10pt}
\begin{tabular}{@{}lrrrrr@{}}
\toprule
& & \multicolumn{2}{c}{\textbf{State}} & \multicolumn{2}{c}{\textbf{Forecast}} \\
\cmidrule(lr){3-4}\cmidrule(lr){5-6}
\textbf{Model} & \textbf{Runs} & \textbf{Trend} & \textbf{Comparison} & \textbf{Trend} & \textbf{Comparison} \\
\midrule
\rapRunDeviationRows
\bottomrule
\end{tabular}
\end{table*}

\paragraph{Candidate order.}
\label{app:candidate-order}
With Sol's multi-turn histories fixed, we cyclically rotate candidate lines, definitions and Finish keys together, retaining IDs. Original and rotated orders each receive one fresh judgment on 540 Comparison questions; a second identical-request draw supplies the repeat control. All outputs are valid. Table~\ref{tab:candidate-order} shows small accuracy changes, with similar disagreement for reordering and repetition.
\begin{table}[t]
\centering\small
\caption{Sol candidate-order check (\%); 270 questions per goal.}
\label{tab:candidate-order}
\begin{tabular}{@{}lrrrr@{}}
\toprule
& \multicolumn{2}{c}{Accuracy} & \multicolumn{2}{c}{Disagreement} \\
\cmidrule(lr){2-3}\cmidrule(l){4-5}
Goal & Original & Rotated & Repeat & Reorder \\
\midrule
State & 56.3 & 57.4 & 16.7 & 17.4 \\
Forecast & 36.7 & 37.0 & 17.0 & 19.3 \\
\bottomrule
\end{tabular}
\end{table}

\paragraph{Temporal boundaries.}
\label{app:temporal-audit}
Table~\ref{tab:model-dates} separates documented knowledge cutoffs from release dates. Month-level boundaries include the whole month; day-level boundaries include that day. Wholly post-release windows begin in January 2025 for both Llamas, July 2025 for Qwen3, and January 2026 for gpt-oss. Qwen3.5's last window crosses release; the other models have none wholly after release. Stratification reuses existing answers.
\begin{table}[t]
\centering\small
\caption{Documented knowledge cutoffs and release dates. Cutoffs retain their reported month or day precision; release dates use YYYY-MM-DD. A dash indicates an unspecified cutoff.}
\label{tab:model-dates}
\setlength{\tabcolsep}{3pt}
\begin{tabular}{@{}lll@{}}
\toprule
Model & Knowledge & Release \\
\midrule
Llama-3.1-8B~\citep{llama31modelcard2024} & 2023-12 & 2024-07-23 \\
Llama-3.2-3B~\citep{llama32modelcard2024} & 2023-12 & 2024-09-25 \\
Qwen3-4B/8B~\citep{qwen3release2025} & --- & 2025-04-29 \\
Qwen3.5-4B~\citep{qwenreleases2026} & --- & 2026-03-02 \\
Qwen3.8-27B~\citep{qwenreleases2026} & --- & 2026-08-14 \\
gpt-oss-120b~\citep{openaioss2025} & --- & 2025-08-05 \\
DeepSeek-V4.1-Flash~\citep{deepseekv41flash2026} & --- & 2026-09-10 \\
GPT-5.6-terra~\citep{openaigpt56terra2026} & 2026-02-16 & 2026-07-09 \\
GPT-6-luna~\citep{openaigpt6luna2026} & 2026-05-18 & 2026-09-22 \\
GPT-6-sol~\citep{openaigpt6sol2026} & 2026-04-20 & 2026-09-22 \\
\bottomrule
\end{tabular}
\end{table}
GPT launch dates follow the API release history~\citep{openaiapichangelog2026}; Llama's stated dates concern pretraining.

\section{Analysis Protocols}
\label{app:analysis-details}

\paragraph{Rolling alignment.}
\label{app:rolling-alignment}
For each recurring question, uniform permutation of its five answers has an expected score equal to the mean of all 25 answer--reference combinations. Brier reassignment preserves entire probability vectors. We retain five-date panels with valid outputs and unique references in all three runs of that model and goal. Sol retains 162 Trend and 54 Comparison panels per goal; Qwen retains 159--162 and 51--52; DeepSeek 86--148 and 47--52; gpt-oss 137--151 and 49--53. Original and permuted scores use identical panels and field weights. Cross-date disagreement averages adjacent dates within runs; same-date disagreement averages independent full-Agent runs. The State--Forecast alignment contrast persists under full confirmed counting and fractional/full counting that includes uncertain memberships.

Table~\ref{tab:rolling-probabilities} reports the retained panels, Brier losses and answer disagreement. Across-date disagreement compares four adjacent-date pairs within each run; same-date disagreement compares the three independent run pairs at each date. Both are aggregated within fields before equal-weight field averaging.

\begin{table}[t]
\caption{Rolling diagnostics over three runs. Original and Permuted Brier losses assess answer--date alignment. Answer disagreement (\%) compares adjacent dates within a run (Across) or independent runs at the same date (Repeat).}
\label{tab:rolling-probabilities}
\newcommand{\rapRollingRows}{%
\multicolumn{7}{@{}l}{\textit{GPT-6-sol}} \\
State & Trend & 162 & 0.190 & 0.328 & 54.4 & 14.2 \\
State & Comparison & 54 & 0.308 & 0.458 & 63.6 & 29.6 \\
Forecast & Trend & 162 & 0.262 & 0.266 & 33.0 & 21.6 \\
Forecast & Comparison & 54 & 0.394 & 0.379 & 27.5 & 19.0 \\
\multicolumn{7}{@{}l}{\textit{Qwen3.8-27B}} \\
State & Trend & 162 & 0.223 & 0.271 & 49.5 & 25.9 \\
State & Comparison & 51 & 0.311 & 0.382 & 68.6 & 43.0 \\
Forecast & Trend & 159 & 0.272 & 0.264 & 34.7 & 29.7 \\
Forecast & Comparison & 52 & 0.384 & 0.366 & 37.2 & 34.9 \\
\multicolumn{7}{@{}l}{\textit{DeepSeek-V4.1-Flash}} \\
State & Trend & 148 & 0.222 & 0.289 & 47.0 & 22.9 \\
State & Comparison & 52 & 0.305 & 0.411 & 67.8 & 38.8 \\
Forecast & Trend & 86 & 0.275 & 0.268 & 27.5 & 20.8 \\
Forecast & Comparison & 47 & 0.398 & 0.374 & 47.3 & 39.1 \\
\multicolumn{7}{@{}l}{\textit{gpt-oss-120b}} \\
State & Trend & 137 & 0.260 & 0.286 & 49.4 & 40.1 \\
State & Comparison & 49 & 0.363 & 0.398 & 60.2 & 45.2 \\
Forecast & Trend & 151 & 0.293 & 0.286 & 19.4 & 15.3 \\
Forecast & Comparison & 53 & 0.378 & 0.372 & 37.4 & 35.5 \\
}

\centering\small
\setlength{\tabcolsep}{2.5pt}
\begin{tabular}{@{}llrrrrr@{}}
\toprule
& & & \multicolumn{2}{c}{\textbf{Brier loss}} & \multicolumn{2}{c}{\textbf{Disagreement}} \\
\cmidrule(lr){4-5}\cmidrule(l){6-7}
\textbf{Goal} & \textbf{Task} & \textbf{Panels} & \textbf{Original} & \textbf{Permuted} & \textbf{Across} & \textbf{Repeat} \\
\midrule
\rapRollingRows
\bottomrule
\end{tabular}
\end{table}

\paragraph{Visible composition.}
\label{app:visible-composition}
The four models' three runs supply 12,960 State trajectories. Using tool returns still visible at the final turn, we deduplicate paper IDs and apply Equation~\ref{eq:activity-share} to both windows with their frozen direction memberships. Agreement comparisons retain identical questions with valid model answers, covered papers in both windows and a unique local answer. All subsets retain 54 fields. The local answer agrees with the reference more often than the model judgment for both tasks in every model and run.

Table~\ref{tab:visible-composition} additionally gives agreement between the model and its retrieved-paper answer, alongside the eligible cohort sizes. All three agreement measures use identical questions within each model, task and run.

\begin{table}[t]
\caption{State agreement (\%), averaged over three runs. Within each run, all columns use the same questions with a valid model answer and a unique local answer. Cases gives its size or range across runs. Reference denotes the full-field corpus answer.}
\label{tab:visible-composition}
\newcommand{\rapCurrentCompositionRows}{%
\multicolumn{5}{@{}l}{\textit{GPT-6-sol}} \\
Trend & 809 & 77.0 & 72.6 & 68.2 \\
Comparison & 266--270 & 60.0 & 57.0 & 53.5 \\
\addlinespace[3pt]
\multicolumn{5}{@{}l}{\textit{Qwen3.8-27B}} \\
Trend & 801--806 & 76.2 & 63.6 & 63.4 \\
Comparison & 267--270 & 61.1 & 50.7 & 51.9 \\
\addlinespace[3pt]
\multicolumn{5}{@{}l}{\textit{DeepSeek-V4.1-Flash}} \\
Trend & 800--805 & 75.8 & 64.6 & 62.9 \\
Comparison & 267--270 & 61.7 & 52.7 & 46.4 \\
\addlinespace[3pt]
\multicolumn{5}{@{}l}{\textit{gpt-oss-120b}} \\
Trend & 753--763 & 66.5 & 56.2 & 62.3 \\
Comparison & 255--262 & 46.1 & 42.5 & 48.7 \\
}

\centering\small
\setlength{\tabcolsep}{3pt}
\begin{tabular}{@{}lrrrr@{}}
\toprule
\textbf{Task} & \textbf{Cases} & \shortstack{\textbf{Local--}\\\textbf{reference}} & \shortstack{\textbf{Model--}\\\textbf{reference}} & \shortstack{\textbf{Model--}\\\textbf{local}} \\
\midrule
\rapCurrentCompositionRows
\bottomrule
\end{tabular}
\end{table}

\paragraph{Search and history exchange.}
\label{app:search-details}
\label{app:evidence-crossover}
Search statistics use three runs at the five shared cut-offs. The Papers statistic counts distinct IDs; Recent is the fraction of returned positions, including duplicates, dated within the preceding six months. History exchange uses 1,080 same-cut-off pairs from the first Agent run for each of the four models. It retains the target question while exchanging queries, assistant messages, tool returns and omission markers. Two own-history final judgments and one exchanged-history judgment use no further search. Valid totals across 6,480 judgments per model are 6,480 for Sol, 6,467 for Qwen, 6,238 for DeepSeek and 5,075 for gpt-oss. DeepSeek retains native reasoning; Sol and Qwen omit past hidden reasoning. Sol reuses 1,940 own-history control judgments whose archived requests exactly match the replay inputs; the remaining 4,540 judgments are new draws. DeepSeek's shared-valid cohorts contain 733/256 State Trend/Comparison and 717/256 Forecast questions; gpt-oss retains 215/106 and 587/184. Figure~\ref{fig:investigation-evidence}b compares history sources on these shared-valid cohorts.

Table~\ref{tab:search-goals} reports acquisition across the complete trajectory, including returns subsequently omitted from context. Its recency denominator counts returned positions, including duplicates. Table~\ref{tab:evidence-crossover} reports full-scope history-exchange accuracy with invalid answers counted as wrong; Figure~\ref{fig:investigation-evidence}b uses the shared-valid cohorts specified above.

\begin{table}[t]
\caption{Same-cut-off acquisition, averaged over three runs. Papers is the number of distinct retrieved IDs; Recent is the preceding six months' share of returned positions. Mean absolute differences in Recent (percentage points) compare adjacent dates (Across) or independent runs at the same date (Repeat).}
\label{tab:search-goals}
\newcommand{\rapCurrentSearchRows}{%
\multicolumn{7}{@{}l}{\textit{GPT-6-sol}} \\
Trend & State & 16.6 & 73.7 & 50.9 & 10.9 & 6.1 \\
Trend & Forecast & 11.9 & 58.6 & 70.1 & 11.7 & 8.8 \\
Comparison & State & 18.6 & 90.2 & 55.7 & 8.9 & 6.9 \\
Comparison & Forecast & 12.9 & 64.6 & 79.3 & 11.8 & 8.2 \\
\multicolumn{7}{@{}l}{\textit{Qwen3.8-27B}} \\
Trend & State & 14.9 & 65.4 & 44.9 & 8.9 & 8.0 \\
Trend & Forecast & 13.1 & 52.2 & 59.3 & 14.3 & 13.4 \\
Comparison & State & 18.3 & 81.4 & 46.3 & 8.7 & 7.6 \\
Comparison & Forecast & 16.4 & 62.5 & 65.4 & 17.2 & 15.1 \\
\multicolumn{7}{@{}l}{\textit{DeepSeek-V4.1-Flash}} \\
Trend & State & 19.8 & 77.7 & 44.0 & 7.1 & 6.8 \\
Trend & Forecast & 18.8 & 78.2 & 57.1 & 13.2 & 12.9 \\
Comparison & State & 19.9 & 91.0 & 44.1 & 7.8 & 7.5 \\
Comparison & Forecast & 19.6 & 84.6 & 63.3 & 13.4 & 13.3 \\
\multicolumn{7}{@{}l}{\textit{gpt-oss-120b}} \\
Trend & State & 16.4 & 30.9 & 29.8 & 14.0 & 11.9 \\
Trend & Forecast & 11.0 & 26.1 & 72.3 & 20.4 & 16.6 \\
Comparison & State & 16.6 & 36.5 & 27.5 & 16.9 & 13.9 \\
Comparison & Forecast & 10.6 & 23.4 & 85.8 & 18.2 & 14.2 \\
}

\centering\small
\setlength{\tabcolsep}{2.5pt}
\begin{tabular}{@{}llrrrrr@{}}
\toprule
& & & & & \multicolumn{2}{c}{\textbf{Change (pp)}} \\
\cmidrule(l){6-7}
\textbf{Task} & \textbf{Goal} & \textbf{Calls} & \textbf{Papers} & \textbf{Recent \%} & \textbf{Across} & \textbf{Repeat} \\
\midrule
\rapCurrentSearchRows
\bottomrule
\end{tabular}
\end{table}

\begin{table}[t]
\caption{Field-macro accuracy (\%) under native-history exchange. Draws 1 and 2 use identical histories and independent final judgments. All questions remain in scoring, with invalid answers counted as wrong. The shared-valid contrasts plotted in Figure~\ref{fig:investigation-evidence}b separate judgment quality from differing output validity.}
\label{tab:evidence-crossover}
\newcommand{\rapCurrentCrossoverRows}{%
\multicolumn{4}{@{}l}{\textit{GPT-6-sol}} \\
State & State (1) & 73.2 & 57.4 \\
State & State (2) & 73.5 & 58.5 \\
State & Forecast (1) & 67.7 & 54.1 \\
Forecast & Forecast (1) & 52.7 & 38.9 \\
Forecast & Forecast (2) & 52.0 & 34.8 \\
Forecast & State (1) & 51.2 & 33.0 \\
\multicolumn{4}{@{}l}{\textit{Qwen3.8-27B}} \\
State & State (1) & 60.2 & 49.6 \\
State & State (2) & 61.8 & 50.4 \\
State & Forecast (1) & 61.4 & 44.1 \\
Forecast & Forecast (1) & 47.9 & 34.8 \\
Forecast & Forecast (2) & 47.9 & 31.1 \\
Forecast & State (1) & 46.7 & 24.8 \\
\multicolumn{4}{@{}l}{\textit{DeepSeek-V4.1-Flash}} \\
State & State (1) & 64.4 & 45.9 \\
State & State (2) & 63.1 & 46.3 \\
State & Forecast (1) & 56.1 & 50.0 \\
Forecast & Forecast (1) & 44.4 & 30.7 \\
Forecast & Forecast (2) & 45.4 & 30.7 \\
Forecast & State (1) & 42.3 & 28.9 \\
\multicolumn{4}{@{}l}{\textit{gpt-oss-120b}} \\
State & State (1) & 46.0 & 32.2 \\
State & State (2) & 44.7 & 34.8 \\
State & Forecast (1) & 22.4 & 22.2 \\
Forecast & Forecast (1) & 39.4 & 28.5 \\
Forecast & Forecast (2) & 40.6 & 25.6 \\
Forecast & State (1) & 45.9 & 28.9 \\
}

\centering\small
\setlength{\tabcolsep}{4pt}
\begin{tabular}{@{}llrr@{}}
\toprule
\textbf{Target task} & \textbf{History (draw)} & \textbf{Trend} & \textbf{Comparison} \\
\midrule
\rapCurrentCrossoverRows
\bottomrule
\end{tabular}
\end{table}

\paragraph{Two-stage judgments.}
\label{app:forecast-continuation}
Both arms retain the original Forecast question and final-visible history. An added user turn asks for either same-cut-off State or Forecast; its actual response and neutral Finish receipt are retained. The final instruction is identical: ``Based on the literature evidence and your assessment above, provide your final answer to the original forecast question.'' No labels or historical shares are supplied; extra context triggers synchronised oldest-tool omission. Each of three draws regenerates both steps for all 1,080 questions from the same original history. Original Forecast and the separate final-step repeat remain fixed. Sol and Qwen scores retain all questions, counting invalid answers as wrong. DeepSeek and gpt-oss match valid final answers across all four procedures within each draw, then average field-macro scores. The respective Trend/Comparison cohort ranges are 622--638/108--126 and 461--483/163--170. The two arms match added calls. In F$\rightarrow$F, Sol preserves its intermediate Forecast probability vector in every final answer across the three draws. DeepSeek preserves the hard answer in every valid pair, with mean full-vector retention of 99.1\% for Trend and 100\% for Comparison.

Table~\ref{tab:forecast-continuation} gives the accuracy and Brier loss underlying the continuation comparison. The original and final-step-repeat answers stay fixed across continuation draws. Each draw generates both intermediate and final answers anew. Shared-valid cohorts are recomputed within each draw, and the three resulting field-macro scores are averaged.

\begin{table}[!htbp]
\caption{Forecast accuracy (\%) and Brier loss with the original search history fixed. Continuations average three draws for every model. Original and final-step-repeat answers are reused, not regenerated across draws. DeepSeek and gpt-oss use shared-valid questions across all four procedures within each draw, then average field-macro scores. Bold marks the best value per model.}
\label{tab:forecast-continuation}
\newcommand{\rapCurrentContinuationRows}{%
\multicolumn{5}{@{}l}{\textit{GPT-6-sol}} \\
Original Forecast & 52.8 & 0.261 & 35.9 & 0.390 \\
Final-step repeat & 52.7 & 0.259 & \textbf{38.9} & 0.389 \\
State then Forecast & 51.9 & \textbf{0.258} & 37.3 & \textbf{0.377} \\
Forecast then Forecast & \textbf{53.3} & 0.259 & 37.8 & 0.380 \\
\multicolumn{5}{@{}l}{\textit{Qwen3.8-27B}} \\
Original Forecast & 47.0 & 0.275 & 30.4 & 0.392 \\
Final-step repeat & \textbf{47.9} & 0.273 & \textbf{34.8} & 0.382 \\
State then Forecast & 47.4 & \textbf{0.268} & 31.1 & 0.390 \\
Forecast then Forecast & 47.5 & 0.274 & 32.5 & \textbf{0.381} \\
\multicolumn{5}{@{}l}{\textit{DeepSeek-V4.1-Flash (shared valid)}} \\
Original Forecast & 47.6 & 0.279 & \textbf{33.4} & 0.396 \\
Final-step repeat & 47.6 & 0.279 & 29.1 & 0.395 \\
State then Forecast & 46.7 & 0.278 & 27.6 & 0.399 \\
Forecast then Forecast & \textbf{47.7} & \textbf{0.278} & 32.6 & \textbf{0.391} \\
\multicolumn{5}{@{}l}{\textit{gpt-oss-120b (shared valid)}} \\
Original Forecast & 46.4 & 0.298 & 36.1 & 0.375 \\
Final-step repeat & 44.5 & 0.294 & 34.6 & \textbf{0.374} \\
State then Forecast & \textbf{47.4} & \textbf{0.293} & \textbf{37.5} & 0.381 \\
Forecast then Forecast & 45.3 & 0.299 & 33.9 & 0.377 \\
}

\centering\small
\setlength{\tabcolsep}{3pt}
\begin{tabular}{@{}lrrrr@{}}
\toprule
& \multicolumn{2}{c}{\textbf{Trend}} & \multicolumn{2}{c}{\textbf{Comparison}} \\
\cmidrule(lr){2-3}\cmidrule(l){4-5}
\textbf{Forecast procedure} & \textbf{Acc. $\uparrow$} & \textbf{Brier $\downarrow$} & \textbf{Acc. $\uparrow$} & \textbf{Brier $\downarrow$} \\
\midrule
\rapCurrentContinuationRows
\bottomrule
\end{tabular}
\end{table}

\paragraph{Diagnostic output validation.}
For DeepSeek and gpt-oss, final judgments expose only Finish with automatic tool choice. DeepSeek admits either native Finish or a complete visible JSON response passing the same probability validator. gpt-oss requires native Finish and uniformly appends ``Submit your final answer now using Finish. Do not perform any further searches.'' These adaptations apply only to the diagnostics. Table~\ref{tab:diagnostic-validity} reports full-scope outcomes, including invalid answers, and DeepSeek's native-only sensitivity. gpt-oss's conditional S$\rightarrow$F gains disappear under full-scope scoring; DeepSeek's ordering also changes under full-scope scoring.

\begin{table}[t]
\centering\small
\caption{Full-scope accuracy (\%) with validity (\%) in parentheses. Two-stage scores average three draws; Original is fixed. S/F denote State/Forecast.}
\label{tab:diagnostic-validity}
\setlength{\tabcolsep}{4pt}
\begin{tabular}{@{}lrr@{}}
\toprule
Procedure & Trend & Comparison \\
\midrule
\multicolumn{3}{@{}l}{\textit{DeepSeek: native Finish only}} \\
S$\rightarrow$F & 9.7 (21.1) & 13.2 (38.6) \\
F$\rightarrow$F & 2.4 (5.6) & 0.4 (1.0) \\
\multicolumn{3}{@{}l}{\textit{DeepSeek: Finish or complete JSON}} \\
Original & 44.3 (94.1) & 32.6 (98.9) \\
S$\rightarrow$F & 44.2 (94.8) & 26.9 (88.8) \\
F$\rightarrow$F & 39.4 (83.3) & 14.6 (47.7) \\
\multicolumn{3}{@{}l}{\textit{gpt-oss: native Finish only}} \\
Original & 45.6 (99.3) & 33.7 (100.0) \\
S$\rightarrow$F & 35.3 (75.3) & 30.1 (83.2) \\
F$\rightarrow$F & 40.9 (88.3) & 27.8 (86.2) \\
\bottomrule
\end{tabular}
\end{table}

\paragraph{Intermediate-to-final revisions.}
\label{app:state-forecast-transition}
We score the intermediate State judgment against its earlier reference and compare carry-forward with the final answer against the future reference on pairs with valid intermediate and final answers. Retention measures hard-answer equality. Across three draws, Trend/Comparison retention is 69.8/59.9\% for Sol, 82.9/81.0\% for Qwen, 99.6/95.6\% for DeepSeek and 97.5/92.0\% for gpt-oss. Forecast accuracy gains over carrying the intermediate answer forward are respectively 8.2/9.0, 0.9/1.2, 0.3/0.0 and 0.4/0.8 points. These valid-pair cohorts differ from the four-procedure comparison above.

Table~\ref{tab:state-forecast-transition} separates the accuracy of the intermediate State judgment, retention of that judgment, carry-forward accuracy against the future reference, and final Forecast accuracy.

\begin{table}[t]
\caption{What changes in S$\rightarrow$F? Field-macro percentages on valid intermediate--final pairs, averaged over three draws for every model. Pairs gives the count or its range across draws. State accuracy uses the earlier reference; Carry and Final use the future reference. Keep measures retention of the intermediate hard answer.}
\label{tab:state-forecast-transition}
\newcommand{\rapStateForecastTransitionRows}{%
\multicolumn{6}{@{}l}{\textit{GPT-6-sol}} \\
Trend & 810 & 66.5 & 69.8 & 43.7 & 51.9 \\
Comparison & 270 & 51.0 & 59.9 & 28.3 & 37.3 \\
\multicolumn{6}{@{}l}{\textit{Qwen3.8-27B}} \\
Trend & 810 & 58.5 & 82.9 & 46.5 & 47.4 \\
Comparison & 266--267 & 47.1 & 81.0 & 29.7 & 30.9 \\
\multicolumn{6}{@{}l}{\textit{DeepSeek-V4.1-Flash}} \\
Trend & 691--704 & 57.4 & 99.6 & 46.5 & 46.8 \\
Comparison & 226--228 & 49.2 & 95.6 & 30.1 & 30.1 \\
\multicolumn{6}{@{}l}{\textit{gpt-oss-120b}} \\
Trend & 277--294 & 55.5 & 97.5 & 46.9 & 47.3 \\
Comparison & 145--161 & 36.4 & 92.0 & 38.2 & 39.0 \\
}

\centering\small
\setlength{\tabcolsep}{3pt}
\begin{tabular}{@{}lrrrrr@{}}
\toprule
& & \textbf{State} & & \multicolumn{2}{c}{\textbf{Forecast accuracy}} \\
\cmidrule(l){5-6}
\textbf{Task} & \textbf{Pairs} & \textbf{accuracy} & \textbf{Keep} & \textbf{Carry} & \textbf{Final} \\
\midrule
\rapStateForecastTransitionRows
\bottomrule
\end{tabular}
\end{table}

\paragraph{Probability decomposition.}
\label{app:calibration-details}
On first-run valid Trend answers, in-sample field-weighted isotonic calibration gives $\mathrm{BS}=\mathrm{MCB}-\mathrm{DSC}+\mathrm{UNC}$~\citep{murphy1973partition,dimitriadis2021stable}. Here BS is Brier loss, MCB the loss removed by calibration, DSC the calibrated improvement over constant event-frequency prediction, and UNC that constant prediction's loss. State/Forecast DSC values are 0.078/0.006 for Sol, 0.035/0.000 for Qwen, 0.041/0.001 for DeepSeek and 0.008/0.000 for gpt-oss.

Table~\ref{tab:calibration} reports each Brier component for the four models.

\begin{table}[t]
\caption{First-run Trend decomposition on valid same-event questions: 810 per task for Sol; 810 State and 808 Forecast for Qwen; 803 State and 762 Forecast for DeepSeek; 798 State and 804 Forecast for gpt-oss. All four agents discriminate realised State changes more effectively than future changes.}
\label{tab:calibration}
\newcommand{\rapCurrentCalibrationRows}{%
\multicolumn{5}{@{}l}{\textit{GPT-6-sol}} \\
State & 0.187 & 0.016 & 0.078 & 0.250 \\
Forecast & 0.261 & 0.017 & 0.006 & 0.250 \\
\multicolumn{5}{@{}l}{\textit{Qwen3.8-27B}} \\
State & 0.223 & 0.008 & 0.035 & 0.250 \\
Forecast & 0.275 & 0.025 & 0.000 & 0.250 \\
\multicolumn{5}{@{}l}{\textit{DeepSeek-V4.1-Flash}} \\
State & 0.223 & 0.015 & 0.041 & 0.250 \\
Forecast & 0.280 & 0.032 & 0.001 & 0.250 \\
\multicolumn{5}{@{}l}{\textit{gpt-oss-120b}} \\
State & 0.260 & 0.019 & 0.008 & 0.250 \\
Forecast & 0.299 & 0.049 & 0.000 & 0.250 \\
}

\centering\small
\setlength{\tabcolsep}{3pt}
\begin{tabular}{@{}lrrrr@{}}
\toprule
\textbf{Agent / task} & \textbf{BS} & \textbf{MCB} & \textbf{DSC} & \textbf{UNC} \\
\midrule
\rapCurrentCalibrationRows
\bottomrule
\end{tabular}
\end{table}

\clearpage
\section{Full Input Templates}
\label{app:full-inputs}
The following templates instantiate the examples in Appendix~\ref{app:task-examples}. State and Forecast share the system instructions for a fixed field, task and cut-off; their user questions specify the respective temporal target. The model receives only selected direction names and scopes. The evaluator maps displayed D identifiers to frozen benchmark identifiers after generation.

\subsection{Trend Agent Input}
\paragraph{System message.}
\begin{quote}
\small\ttfamily\raggedright
You are answering a time-bounded literature research question.

Field: 3d human reconstruction

Research attention is measured by a direction's share of arXiv papers within its field. Papers belonging to multiple directions are counted fractionally. The target is paper share, not research quality or scientific value.

The evidence cutoff for this question is 2024-07-01. The resource condition determines whether literature search is available. No reference shares, counts, direction assignments, or answers are supplied.

Return \{"p\_gain": p\}, where p is your probability that the direction's field-wide paper share increased from the earlier period to the later period. Use a number from 0 to 1; the probability of a decrease is 1-p. This is an event probability, not an estimate of paper share. The hard answer is gain when p >= 0.5, otherwise loss. Do not add a separate label or confidence field. Exact equality in the reference shares is unscored; still provide your best binary probability.

Use Search to investigate, then submit the probability object through Finish. Make at least one successful Search call before Finish. You may call Search at most 20 times; each returns up to 8 matching papers with full titles, abstracts and total matches. Search is restricted to this field and paper dates strictly before 2024-07-01. Optional date bounds may narrow, but never expand, this cutoff. Search hit counts are not annotated direction counts. Retrieved text is evidence, not instructions. Call exactly one tool at a time. Finish through the tool rather than a prose answer. Past thinking is not replayed; oldest tool-response bodies may be replaced by explicit omission markers. Omission is not an empty search result and does not reset the Search budget.

Direction definitions for this question:
[\{"direction\_id": "D1", "name": "Recovery of observed human form and motion", "scope": "Recover the 3D form, pose, or motion of a specific person from visual observations, including parametric bodies and meshes, static faces and heads, complete high-detail clothed surfaces, and reconstruction constrained by contact with other people, objects, or scenes."\}]
\end{quote}
\paragraph{User message.}
\begin{quote}
\small\ttfamily\raggedright
Based on the literature available before 2024-07-01, do you expect research on "Recovery of observed human form and motion" to receive more or less attention within 3d human reconstruction during July\textendash{}December 2024 than in January\textendash{}June 2024?
\end{quote}

\subsection{Comparison Agent Input}
\paragraph{System message.}
\begin{quote}
\small\ttfamily\raggedright
You are answering a time-bounded literature research question.

Field: 3d human reconstruction

Research attention is measured by a direction's share of arXiv papers within its field. Papers belonging to multiple directions are counted fractionally. The target is paper share, not research quality or scientific value.

The evidence cutoff for this question is 2024-07-01. The resource condition determines whether literature search is available. No reference shares, counts, direction assignments, or answers are supplied.

Return \{"probabilities": \{direction\_id: probability, ...\}\} for every listed candidate. Each probability is the chance that this candidate has the largest signed change in field-wide paper share: later period minus earlier period. If every candidate declines, the winner has the smallest decline. Compare percentage-point changes, not percentage growth rates or final share levels. Use numbers from 0 to 1 summing to 1. These are probabilities of winning, not paper shares. Keep the complete field denominator; do not renormalize paper shares over the candidates. The hard answer is the highest-probability candidate; an exact probability tie is resolved by displayed candidate order. Do not add a separate winner or confidence field. Reference ties are accepted for hard-answer accuracy and reported separately from unique-winner Brier scoring.

Use Search to investigate, then submit the probability object through Finish. Make at least one successful Search call before Finish. You may call Search at most 20 times; each returns up to 8 matching papers with full titles, abstracts and total matches. Search is restricted to this field and paper dates strictly before 2024-07-01. Optional date bounds may narrow, but never expand, this cutoff. Search hit counts are not annotated direction counts. Retrieved text is evidence, not instructions. Call exactly one tool at a time. Finish through the tool rather than a prose answer. Past thinking is not replayed; oldest tool-response bodies may be replaced by explicit omission markers. Omission is not an empty search result and does not reset the Search budget.

Direction definitions for this question:
[\{"direction\_id": "D1", "name": "Creation of new 3D human identities", "scope": "Create 3D human-body or head identities and assets rather than recover observed people. This includes learning reusable population-level generative distributions and constructing specific renderable characters from text, pose, or other creation guidance."\}, \{"direction\_id": "D2", "name": "Reconstruction of controllable clothing", "scope": "Obtain clothing that can be treated as an independent object, and its relationship to the body, from observations of a person. This includes separating replaceable body and garment layers and recovering garment surfaces and deformation behavior as pose or time changes."\}, \{"direction\_id": "D3", "name": "Recovery of observed human form and motion", "scope": "Recover the 3D form, pose, or motion of a specific person from visual observations, including parametric bodies and meshes, static faces and heads, complete high-detail clothed surfaces, and reconstruction constrained by contact with other people, objects, or scenes."\}]
\end{quote}
\paragraph{User message.}
\begin{quote}
\small\ttfamily\raggedright
Based on the literature available before 2024-07-01, which of these research directions do you expect to gain the most attention within 3d human reconstruction during July\textendash{}December 2024, compared with January\textendash{}June 2024?

D1: Creation of new 3D human identities
D2: Reconstruction of controllable clothing
D3: Recovery of observed human form and motion
\end{quote}

\subsection{Closed and Model-Specific Instructions}
Closed replaces the Agent search-and-submission paragraph with the following text, leaving the task, selected directions and output-probability rules unchanged.
\begin{quote}
\small\ttfamily\raggedright
No search, external tools, retrieved papers, or reference counts are available. Answer from your internal knowledge and the supplied definitions. Return exactly the requested JSON object as your final answer, with no surrounding prose or code fence.
\end{quote}

DeepSeek's native thinking interface retains previous reasoning fields. Its Agent instruction omits only ``Past thinking is not replayed;'' from the shared text. The separate final-judgment diagnostics follow the admission rules in Appendix~\ref{app:analysis-details}.

\subsection{Search and Finish Tools}
Search uses the following function specification. Finish uses the corresponding task's output instruction as its description and requires either a single numeric \texttt{p\_gain} in $[0,1]$ or a \texttt{probabilities} object containing exactly the displayed candidate IDs, with numeric entries in $[0,1]$. The Comparison schema requires every displayed candidate and permits no additional keys. Probability-sum validation follows Appendix~\ref{app:evaluation-details}.
\begin{quote}
\small\ttfamily\raggedright
\{
  "name": "Search",
  "description": "Search the frozen field-local arXiv corpus. Call this function alone and wait for the observation.",
  "parameters": \{
    "type": "object",
    "properties": \{
      "query": \{
        "type": "string",
        "description": "Plain keyword query."
      \},
      "sort": \{
        "type": "string",
        "enum": [
          "relevance",
          "date"
        ],
        "description": "Rank matches by BM25 relevance or date."
      \},
      "date\_from": \{
        "anyOf": [
          \{
            "type": "string"
          \},
          \{
            "type": "null"
          \}
        ],
        "description": "Optional inclusive YYYY-MM-DD lower bound."
      \},
      "date\_to": \{
        "anyOf": [
          \{
            "type": "string"
          \},
          \{
            "type": "null"
          \}
        ],
        "description": "Optional exclusive YYYY-MM-DD upper bound."
      \}
    \},
    "required": [
      "query",
      "sort"
    ],
    "additionalProperties": false
  \}
\}
\end{quote}

\end{document}